%% file: main.tex
\documentclass[letterpaper]{article} 
\usepackage{aaai24}  

\usepackage{amssymb}
\usepackage{pifont} 

\usepackage{times}  
\usepackage{helvet}  
\usepackage{courier}  
\usepackage[hyphens]{url}  
\usepackage{graphicx} 
\usepackage{natbib}
\usepackage{caption}
\usepackage{amstext}
\usepackage{varwidth}
\usepackage{tcolorbox}
\usepackage{multirow}

\newcommand{\cmark}{\ding{51}}%
\newcommand{\xmark}{\ding{55}}%

\newcommand{\datasetFont}{\text}

\newcommand{\ours}{\datasetFont{POPR}\xspace}
\newcommand{\oursabc}{\datasetFont{POPR-EABC}\xspace}

\usepackage{microtype}
\usepackage{subfigure}
\usepackage{booktabs} 
\usepackage{graphicx}
\usepackage{booktabs}
\usepackage{amsmath}
\usepackage{amssymb}
\usepackage{color}
\usepackage{longtable}
\usepackage{etoolbox}

\usepackage{dsfont}
\usepackage[font=small,labelfont=bf]{caption}
\usepackage{amsmath, amsfonts, amsthm, amssymb}
\usepackage[lined,boxed,commentsnumbered,ruled,vlined]{algorithm2e}

\usepackage{mathtools}
\usepackage{color}
\usepackage{adjustbox}
\usepackage{floatflt}
\usepackage{lipsum} 

\newtheorem{definition}{Definition}

\newcommand{\agr}{\datasetFont{AgreeRank}\xspace}

\newcommand{\fqe}{\datasetFont{FQE}\xspace}
\newcommand{\tree}{\datasetFont{Tree-Backup ($\lambda$)}\xspace}
\newcommand{\Qpi}{\datasetFont{$Q^\pi(\lambda)$}\xspace}
\newcommand{\MBased}{\datasetFont{MBased}\xspace}
\newcommand{\dice}{\datasetFont{BayesDICE}\xspace}

\newcommand{\EnvFont}{\texttt}
\newcommand{\toy}{\EnvFont{ToyEnv}\xspace}
\newcommand{\mcd}{\EnvFont{MountainCar}\xspace}

\newcommand{\acro}{\EnvFont{AcroBot}\xspace}
\newcommand{\pend}{\EnvFont{Pendulumn}\xspace}
\newcommand{\bip}{\EnvFont{BipedalWalker}\xspace}

\usepackage{natbib}  
\usepackage{caption} 
\title{Probabilistic Offline Policy Ranking with Approximate Bayesian Computation}
\author{
    Longchao Da\textsuperscript{\rm 1}, Porter Jenkins\textsuperscript{\rm 2}, Trevor Schwantes\textsuperscript{\rm 2}, Jeffrey Dotson\textsuperscript{\rm 2}, Hua Wei\textsuperscript{\rm 1}\thanks{Corresponding Author}
}
\affiliations{
    \textsuperscript{\rm 1}Arizona State University, 
    \textsuperscript{\rm 2}Brigham Young University\\

    \{longchao,hua.wei\}@asu.edu, pjenkins@cs.byu.edu, Schwantes2@gmail.com, jeff\_dotson@byu.edu
}

\usepackage{bibentry}

\begin{document}

\maketitle

\begin{abstract}
In practice, it is essential to compare and rank candidate policies offline before real-world deployment for safety and reliability. Prior work seeks to solve this offline policy ranking (OPR) problem through value-based methods, such as Off-policy evaluation (OPE). However, they fail to analyze special case performance (e.g., worst or best cases), due to the lack of holistic characterization of policies’ performance. It is even more difficult to estimate precise policy values when the reward is not fully accessible under sparse settings.
In this paper, we present Probabilistic Offline Policy Ranking (\ours), a framework to address OPR problems by leveraging expert data to characterize the probability of a candidate policy behaving like experts, and approximating its entire performance posterior distribution to help with ranking. \ours does not rely on value estimation, and the derived performance posterior can be used to distinguish candidates in worst-, best-, and average-cases. To estimate the posterior, we propose \oursabc, an Energy-based Approximate Bayesian Computation (ABC) method conducting likelihood-free inference. \oursabc reduces the heuristic nature of ABC by a smooth energy function, and improves the sampling efficiency by a pseudo-likelihood. We empirically demonstrate that \oursabc is adequate for evaluating policies in both discrete and continuous action spaces across various experiment environments, and facilitates probabilistic comparisons of candidate policies before deployment.

\end{abstract}

\input{intro}

\input{preliminaries}

\input{method_hua}

\input{experiments}

\input{relatedwork}

\input{conclusion}

\bibliography{aaai24}

\clearpage
\newpage
\input{Appendix}

\end{document}

%% file: intro.tex
\section{Introduction}

Policies instruct actions in decision making. With a set of candidate policies, evaluating and ranking prior to real deployment is critical for real-world applications. Off-policy evaluation (OPE) allows one to estimate the goodness of a policy (often referred to as target/candidate policy) using data collected from another, possibly unrelated policy (referred to as behavior policy). 
Such evaluation is important because testing or implementing a policy that performs poorly can be costly (e.g., in trading or traffic control) or even potentially harmful (e.g., in drug trials or autonomous vehicles).


With growing interest in OPE, the research community has produced a number of estimators, including importance sampling (IS)~\cite{Hammersley1964MonteCM,Powell1966WeightedUS,Horvitz1952AGO,Precup2000EligibilityTF,Thomas2016DataEfficientOP,Farajtabar2018MoreRD,Dudk2011DoublyRP, Jiang2015DoublyRO},  direct methods (DM)~\cite{Harutyunyan2016QWO,Li2010UnbiasedOE,Le2019BatchPL,Kostrikov2020StatisticalBF} and distribution correction estimation (DICE) methods~\cite{Dai2020GenDICEOG,Zhang2020GradientDICERG,Yang2020OffPolicyEV,Yang2020OfflinePS}. Importance-sampling-based methods weight the data collected from the behavior policy according to the probability of transitioning to each state under the target policy, yet they assume access to a probability distribution over actions from the behavior policy. DM and DICE do not require knowing the output probabilities of the behavior policy, where DM directly learns an environment or value model from offline data, and DICE methods learn to estimate the discounted stationary distribution ratios. Most of these methods compute the point estimates of the policy's value~\cite{Dudk2011DoublyRP,Jiang2015DoublyRO,Zhang2020GradientDICERG,Yang2020OffPolicyEV}, some of which additionally estimate the value with confidence intervals~\cite{Thomas2015HighConfidenceOE,Kuzborskij2020ConfidentOE,Feng2020AccountableOE,Dai2020GenDICEOG,Kostrikov2020StatisticalBF}.

While various estimators have been proposed for off-policy evaluation, in many cases, precise policy value estimation is not necessary. Instead, practitioners often place greater emphasis on the correctness of comparison and ranking of candidate policies. Existing work in Supervised Off-Policy Ranking (SOPR)~\cite{jin2021supervised} takes a supervised learning approach to policy ranking, and requires an adequate training set or policies with explicitly labeled performance. In practice, this approach is challenging because (1) actual policy data is typically limited, and (2) access to labeled performance data is a strict assumption. The behavior policy is usually inaccessible and behavioral data is usually restricted, such as in healthcare or confidential financial trading domains. Additionally, when the policies are hard to differentiate from mean performance, we might care about the performance under special situations like the worst or best cases \cite{agarwal2021deep}, which none of the above literature could address. 

\begin{figure*}[t!]
    \centering
    \includegraphics[width=0.9\textwidth]{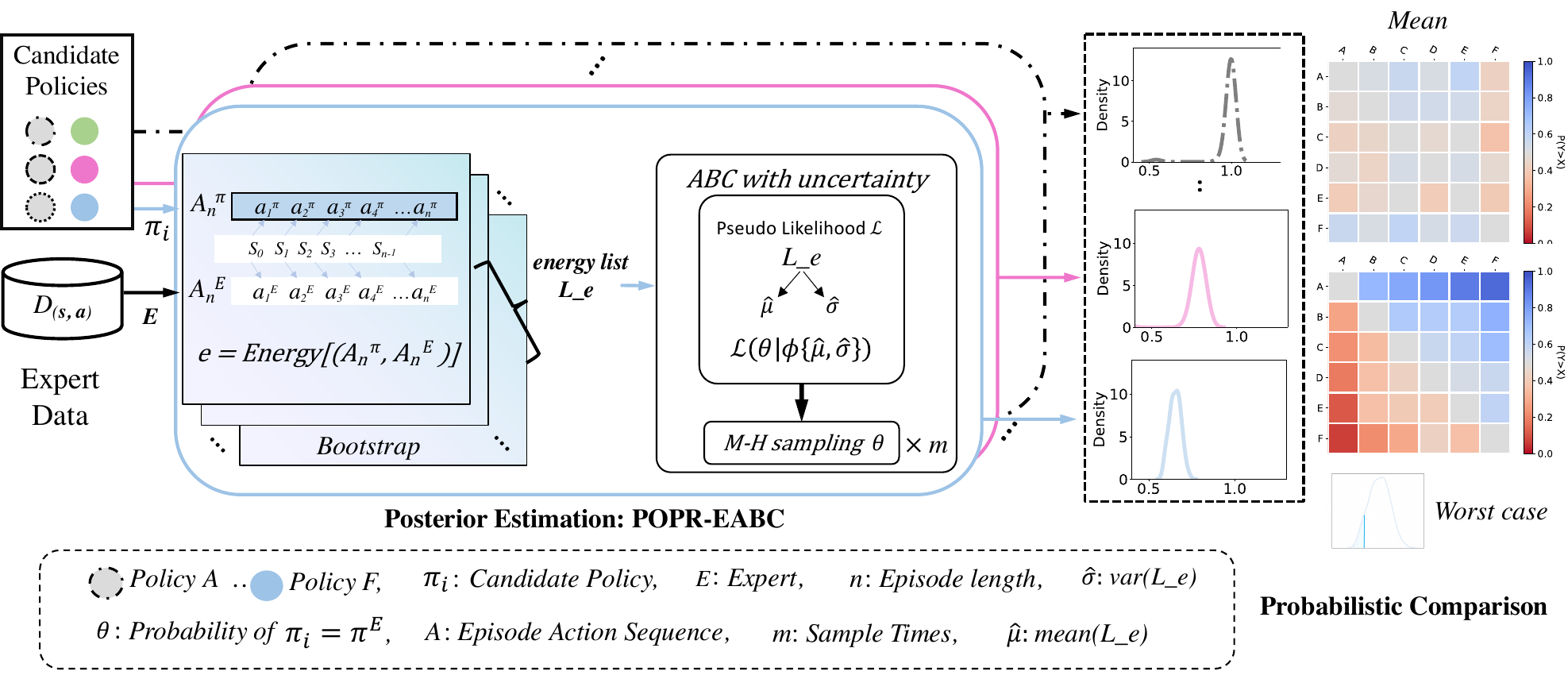}
    \caption{\ours consists of two main parts, posterior estimation and probabilistic comparison. (Left) Given an expert dataset and a set of candidate policies, (Middle) \ours intends to learn the performance posterior distribution of each candidate relative to an expert, in which we propose \oursabc. By bootstrapping, the energy function first calculates a list of energy values between expert action $A_{n}^{E}$ and policy $\pi$'s action $A_{n}^{\pi}$ when given the same $S$ sampled from $D_{(s,a)}$, and then a pseudo-likelihood (aware of mean and variance from energy values) is used in M-H algorithm to sample the acceptable probability $\theta$ from a proposal distribution.  (Right) The sampled proposals approximate the performance posterior distribution, enabling special case comparison, such as worst-case analysis (shown). The heat map is the example pair-wise comparison of the mean (hard to compare) and worst case (clear to distinguish), which shows the benefit of \oursabc. }
    \vspace{-5mm}
    \label{fig:intro}
\end{figure*}

In this paper, we propose the Probabilistic Offline Policy Ranking (\ours) framework to address the above challenges. POPR does not require access to behavioral policies, nor the performance value or reward. Instead, it exploits limited expert data to the maximum extent. Our intuition is that we can measure the expected behavior of a candidate policy relative to a static policy generated by an expert, such as a medical doctor or a driving instructor\cite{kim2013learning}. The more the candidate policy behaves like experts, the better. Based on this, the performance posterior distribution is estimated, providing a holistic characterization of the policy's quality, making it possible to compare the performance in the best or worst case.

Following this framework, we propose a novel method \oursabc, an \underline{\textbf{E}}nergy  based  \underline{\textbf{A}}pproximate  \underline{\textbf{B}}ayesian  \underline{\textbf{C}}omputation. By using a smooth energy function to measure the similarity between expert and policy-generated data, we obviate the need to specify tolerance parameters on summary statistics and improve the efficiency of ABC. We introduce a pseudo-likelihood that parameterizes the energy variance, and facilitates Bayesian inference.


On both self-trained policies and open-sourced policies, we perform extensive evaluations comparing six baselines under different RL tasks covering both discrete and continuous action spaces. The results prove the effectiveness of \oursabc in offline policy evaluation. 
We also demonstrate our method could exploit efficiently at small size expert data and has a high tolerance for data quality.

%% file: preliminaries.tex
\section{Offline Policy Ranking}
In this section, we formalize the Offline Policy Ranking (OPR) problem and the general process of OPR methods. Then we provide a discussion on various solutions to OPR problems. 

\subsection{Formalization of OPR Problem}
We consider a Markov Decision Process (MDP), defined by a tuple ($ S, A, T, R, \gamma $), where $S$ and $A$ represent the state and action spaces, respectively. $ T(s' | s, a)$ represents a, possibly unknown, transition function, where $s'$ is the next possible state from $s$ taking action $a$, $s \in S$, $a \in A$ and $R(s, a)$ represents a reward function. The expected return of a policy $\pi$ is defined as $V(\pi)=\mathbb{E}[\sum^\mathcal{H}_{t=0}\gamma^tr_t]$, where $\mathcal{H}$ is the horizon of the MDP, and $t$ is the index of steps.

\begin{definition} [Offline Policy Ranking]
Given an offline dataset $\mathcal{D}$, that consists of $N$ observed behavior trajectories $\mathcal{T}$ from behavior policy $\mu$. $\mathcal{D} = \{ \mathcal{T}_{i} \langle s_t, a_t, r_t, s_{t+1} \rangle \}_{i=1}^{N}$ with $N$ trajectories, each having a variable length $L_i$, $t\in L_i$. And given a set of candidate policies $ \hat{\Pi} = \{\hat{\pi}^{(1)}, ..., \hat{\pi}^{(k)} \}$, where $k$ represents the policies' index. The goal of offline policy ranking is to acquire a ranked order $O\{\cdot\}$ that represents the true performance of the policies without interacting with the environment online. In later section, OPR stands for Offline Policy Ranking.\\
\label{def1}
\end{definition}
\vspace{-3mm} 

\vspace{-3mm} 
\subsection{Solutions to OPR}
We briefly provide a discussion and comparison of the current solutions to OPR problems, including OPE methods and Supervised Off-Policy Ranking.
\paragraph{OPE Methods}\label{OPE:conclusion} It is possible to solve OPR tasks using Off-policy evaluation by calculating the expected values $\mathbb{E}[V(\hat{\pi}^{(k)}|\mathcal{D})]$ for each candidate 
 policy $\hat{\pi}^{(k)}$ and ranking each $\hat{\pi}^{(k)} \in \hat{\Pi}$ accordingly to obtain $O \{\hat{\Pi} \}$. These methods~\cite{Voloshin2019EmpiricalSO, Harutyunyan2016QWO, precup2000eligibility} aim to precisely estimate the expected value with an offline dataset $\mathcal{D}$, but they either tend to re-weight~\cite{xie2019towards} or provide correction to the original reward values $r$~\cite{nachum2019dualdice} (The discussion on this conclusion can be found in the Appendix~\ref{sec:reward:proof}), such reliance on high-quality and dense value return is a severe challenge for most of OPE methods in practical use. For scenarios with intrinsic sparse reward settings (only receive reward signal when the task is done) or low-quality reward representation (partially observable $(s, a)$ leads to untrackable reward $r$), the OPE methods are likely to rank the policies incorrectly. 
\paragraph{Supervised Off-Policy Ranking (SOPR)}
SOPR~\cite{jin2022supervised} is also able to solve OPR problems by training a ranking model using a policy dataset with labeled performance, and then minimize a ranking loss. It considers the overall performance in the training process but also fails to present a probabilistic result of the comparison, and is not able to conduct worst/best case analysis either. 
We summarize the current solutions in Table~\ref{tab:comparison}, while some of the OPE methods relax the requirement of the action probability $P(a)$ from behavior policy, non of them could alleviate the reliance on rewards, similarly for the SOPR method that relies on true performance for training.


\vspace{-2mm} 
\begin{table}[htbp]
    \centering
        \caption{Comparison of different solutions for OPR tasks}
    \label{tab:comparison}
    \setlength{\tabcolsep}{5mm}{
    \vspace{-2mm} 
    \begin{adjustbox}{center,width=0.32\textwidth}
   \begin{tabular}{ccccc}
    \toprule
    \makebox[0.1\textwidth][c]{Features} & \makebox[0.1\textwidth][c]{OPE} & \makebox[0.1\textwidth][c]{SOPR} & \makebox[0.1\textwidth][c]{POPR-EABC} \\
    \midrule
    Not Access Value & \xmark & \xmark & \cmark \\ 
    Not Access $P(a)$ & \xmark/\cmark & \cmark & \cmark\\ 
    Probabilistic Ranking & \xmark & \xmark & \cmark \\ 
    Case Analysis  & \xmark & \xmark & \cmark \\
    \bottomrule
\end{tabular}
\end{adjustbox}
}
\end{table}
\vspace{-3mm} 

When two policies perform similarly with mean performance, we may wonder what is the probability of one outperforming the other and also the comparison in the worst/best cases behavior, which leads to the probabilistic comparison and case analysis, while none of the existing work could solve. Therefore, we provide our probabilistic framework in Section~\ref{sec:method}. to estimate the posterior distribution of the performance for policy $\hat{\pi}^{(k)}$ on the offline data $\mathcal{D}$. The posterior should contain all the necessary probabilistic information about $\hat{\pi}^{(k)}$, helping us to analyze the best or worst-case performance. In the implementation, we introduce an energy-based inference method with pseudo-likelihood to estimate the posterior, which helps the evaluation to better consider the intrinsic uncertainty.


%% file: method_hua.tex
\section{Probabilistic Offline Policy Ranking}\label{sec:method}

\paragraph{Probabilistic Comparison with Expert Data}
In order to correctly rank the candidate policies, we can compare the performance of the candidate policies with experts as an indicator of the goodness of their performance. Below, we define a statistic value representation $\theta$, which is the result of an estimation, that can be used to rank over $\hat{\Pi}$, and  we provide a formal definition as below.

\begin{definition}[Probabilistic Offline Policy Comparison]\label{sec:ranking}
 Given an expert dataset $\mathcal{D}_e = \{ \langle s_t^{(i)}, a_t^{(i)}, r_t^{(i)}, s_{t+1}^{(i)} \rangle_i \}_{i=1}^{N}$, if we define a statistic $\theta^{(k)} \in [0, 1]$, that measures how consistent the candidate policy $\hat{\pi}^{(k)}$ is with the expert policy, the posterior distribution of $\theta^{(k)}$ is defined as:
 \vspace{-1mm}
\begin{align}\label{def:theta}
    p(\theta^{(k)} | \mathcal{D}_e, \hat{\pi}^{(k)}) \vcentcolon=  p[\pi_{e}(s_t) = \hat{\pi}^{(k)}(s_t) |\  \mathcal{D}_e]
\end{align} 
Note that posterior $p(\theta^{(k)} | \mathcal{D}_e, \hat{\pi}^{(k)})$ can be recognized as the formation of a bag of $\theta ^{(k)}$ samples using certain estimation techniques. 
\end{definition}

Under this definition, $\theta^{(k)} $ is the probability that the candidate policy $\hat{\pi}^{(k)}$ produces the same behavior as the expert $\pi_{e}$, given the state $s_t$. To reduce notational clutter, we use $p(\theta^{(k)} | \mathcal{D}_e, \hat{\pi}^{(k)})$ and $p(\theta^{(k)} | \mathcal{D}_e)$ interchangeably in this paper, additionally, we interchange $\pi_{e}(s_t)$ and $a_e$ since we do not assume to know the form of the expert policy. We will express target policy and behavior policy in OPE scope, whereas candidate policy and expert policy are in the POPR scope.

\paragraph{Posterior Estimation}
Due to the limited number of expert trajectories, real-world environments' stochasticity, and the target policy's decision variance, there is intrinsic uncertainty in $\theta^{(k)} $ when describing the performance of a policy, which introduces bias to the statistic measurement, further causing unreliable evaluation. Holistic depictions considering the variance help to better policies' performance, so we seek to estimate the holistic posterior distribution of $\theta^{(k)} $ for:
\vspace{-1mm}
\begin{equation}\label{eq:posterior}
    p(\theta^{(k)} | \mathcal{D}_e) \propto p(\mathcal{D}_e | \theta^{(k)})p(\theta^{(k)})
\end{equation}
\vspace{-1mm}
Based on the meaning of $\theta^{(k)}$, the posterior $p(\theta^{(k)} | \mathcal{D}_e$) can provide holistic information on policy $\hat{\pi}^{(k)}$'s behavior, and thus supports the evaluation of the candidate policies. The posterior estimation approaches are not limited, such as Bayesian Inference, Markov Chain Monte Carlo, etc. If we notate the posterior process as $f(\cdot)$, we could represent the framework process as below:
\vspace{-1mm}
\begin{equation}
    O\{\hat{\Pi}\} = G(f(\hat{\pi}^{(k)}|\mathcal{T}_i \sim\mathcal{D}_e))
\end{equation}
where $\mathcal{T}_i \in \mathcal{D}_e$ containing $n$ trajectories, $f(\cdot)$ is the posterior estimation process based on sampling trajectories $\mathcal{T}$ from dataset $\mathcal{D}_e$, please note that practitioner could sample multiple times not limited to the total amount $n$ of trajectories. And function $G(\cdot)$ could be any post-process and analysis procedures on posterior samples $\mathcal{S}_{\theta}^{(k)}$, such as statistical calculation or comparison introduced in Section~\ref{sec:abcmethod}. $\mathcal{S}_{\theta}^{(k)}$ represents a bag of sampled $\theta$ for candidate policy $\hat{\pi}^{(k)}$.


\paragraph{Scoring Functions for Ranking and Comparison}\label{method:ranking}

From $f(\cdot)$ process, the derived posterior samples $\mathcal{S}_{\theta}^{(k)}$ summarize all of the performance information learned from the behavior of candidate policies $\hat{\pi}^{(k)}$. We can conduct various tasks such as policy ranking or probabilistic pair-wise comparison, and furthermore, the special cases analysis from the posterior samples. The different tasks will lead to different instantiations of a concrete function $G(\cdot)$.     

$\bullet$~\textit{Ranking on Average:}\label{methiod:ranking}  To conduct a ranking task considering the overall performance, we could use the mean of the whole samples, we do: $\mathbb{E}[\theta^{(k)}] = \frac{1}{|\mathcal{S}|}\sum_{s \in \mathcal{S}_{\theta}^{(k)}} s$, where $|\mathcal{S}|$ stands for the total amount of $\theta^{(k)}$ samples, and then sort $ \hat{\Pi} = \{\hat{\pi}^{(1)}, ..., \hat{\pi}^{(k)} \}$ by $\mathbb{E}[\theta^{(k)}]$, we get the resulting $O\{\hat{\Pi}\}$.

$\bullet$~\textit{Worst/Best-case Analysis:}\label{methiod:case_analysis} To conduct special cases analysis, either ranking or pair-wise comparison, we only need to conduct one step selection on the pre-ordered $\mathcal{S}_{\theta}^{(k)}$. For example, in this paper, when it comes to worst-case comparison, only the bottom  $5\% \times|\mathcal{S}|$ of samples will be selected to keep with the above two analysis procedures. Note that the proportion of observation could vary according to the necessity.

$\bullet$~\textit{Pair-wise Comparison:}\label{methiod:pair-wise} To conduct a pair-wise comparison between a group of policies, we compare by the expected Monte Carlo samples ($\theta$s), e.g., $\hat{\pi}^{(k)}$, and $\hat{\pi}^{(l)}$ by computing $p(\theta^{(k)} > \theta^{(l)} | \mathcal{D}_e)$. In other words, $p(\theta^{(k)} > \theta^{(l)} | \mathcal{D}_e) = \frac{1}{|\mathcal{S}|} \sum_{{s \in \mathcal{S}_{\theta}^{(k)}} } \mathds{1}(\mathcal{S}_{\theta}^{(k)}[i] > \mathcal{S}_{\theta}^{(l)}[i])$, $i$ is the index for each $\theta \in \mathcal{S}_{\theta}^{(k)}$.


\section{\oursabc}\label{sec:abcmethod}
Following the \ours framework, the primary task is to estimate the posterior of a candidate policy $\hat{\pi}^{(k)}$'s performance, in other words, to derive $\mathcal{S}_{\theta}^{(k)}$.
Bayesian inference typically requires the specification of a likelihood function.
In most cases, the data are assumed to be independent and identically distributed (iid) to make the likelihood computation tractable. However, such an assumption is not suitable for policy evaluations since the observed states and actions in MDP's are determined by the environment dynamics, are \textit{not} independent from each other. Thus, we seek to leverage Approximate Bayesian Computation (ABC), which relies on simulation rather than likelihood, to measure the relevance of parameters to the data, the overview of our method is shown in Figure~\ref{fig:intro}.





\subsection{Energy-based Similarity}\label{sec:energy_fun}
The standard ABC paradigm faces the difficulty of designing a good summary statistic for efficient sampling. In practice, these methods suffer from very low acceptance rates and long sampling times \cite{turner2012tutorial}. To overcome this challenge, we define a continuous, scalar-valued energy function, $e = E(\mathcal{T}_e, \hat{\mathcal{T}}), e \in [0, 1]$ to avoid specifying a heuristic discrepancy statistic, where $\mathcal{T}_e$ is the expert trajectory and $\mathcal{\hat{T}}$ is the simulated trajectory taken by candidate policy $\pi^{(k)}$, given the same observation sequence $S_e$. We draw bootstrapped samples of trajectories from our dataset, $d_e \in \mathcal{D}$, and generate simulated data from our candidate policy, $\hat{d}$.  We then calculate the normalized energy between these two data subsets, $E(d_e, \hat{d}) = 1 -\frac{\rho(d_e, \hat{d})}{Z}$, where $Z = |d_e|$  is a normalizing constant, and $\rho$ is could be any universal distance metric. Intuitively, when the similarity between the two bootstrapped data sets is high, $E$ approaches unity; when similarity is low, $E$ approaches zero. In our experiments, we find out that Jensen–Shannon (JS) divergence~\cite{Endres2003ANM} is more efficient compared to other similarity measures, and use it as the default setting. More results w.r.t. different choices of $\rho$ can be found in the Appendix~\ref{app:distance}.



\begin{figure}[t!]
        \centering
     \includegraphics[width=0.25\textwidth]{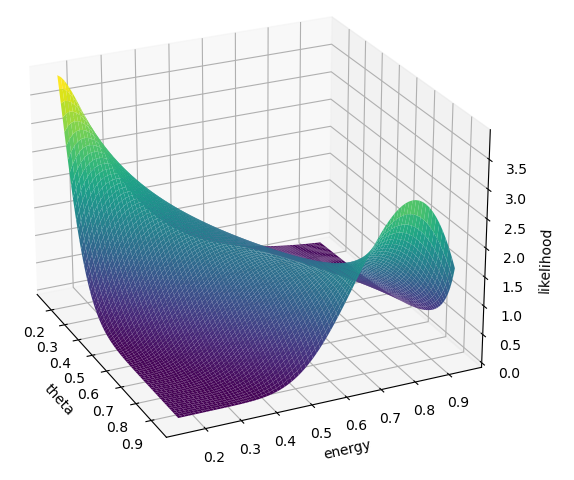}
        \caption{A visual depiction of an example pseudo-likelihood surface. For a given $\theta$, energy tuple, the likelihood of the combination is shown on the vertical axis. For high energy values (indicating high agreement between expert and candidate), large $\theta$ will have a high likelihood (indicating a high acceptance probability). We incorporate this into the ABC-MH (Metropolis-Hastings) sampling algorithm to learn the posterior $p(\theta^{(k)} | \mathcal{D}_e)$ of policy performance.          }
        \label{fig:abc}
        \vspace{-5mm}
\end{figure}

\subsection{Calculating the Pseudo-likelihood} The above energy-based statistic helps mitigate the ABC algorithm's heuristic nature by providing a smooth measure of similarity between datasets. In order to estimate the posterior distribution we design a pseudo-likelihood, which uses the bootstrapped energy values to provide an approximation of the joint probability of the data in a computationally simpler way \cite{besag1975statistical}. Rather than specify a formal likelihood, we fit a density function to the empirical energy values. This density contains distributional information about the behavior of the candidate policy relative to the expert. The pseudo-likelihood, along with the prior, facilitates estimation of the posterior.

More formally, we approximate the likelihood as function of $M$ bootstrapped energy values: $\mathcal{L}( \mathcal{D}_e | \theta^{(k)}) \approx p(\theta^{(k)} | \{e_1, ... e_M\} )$. The energy values, ${e_1, ... e_M}$ are calculated by drawing $M$ bootstrapped datasets $\{ \bar{\mathcal{D}}_1, ... \bar{\mathcal{D}}_M\}$ from the expert data, $\mathcal{D}_e$. This bootstrapping routine induces diversity in both the expert data, and the candidate policy, and facilitates an estimate of variability in $\theta$. We assume that the pseudo-likelihood follows a beta distribution, $p(\theta | \{e_1, ... e_M\}) \sim \textrm{Beta}(\alpha, \beta)$, since the support for the beta distribution lies in $[0, 1]$ and is conducive for estimating probabilities. Because the bootstrapped energy values are all sampled independently, we can use the relatively simple method of moments estimator \cite{fielitz1975concepts} to fit the pseudo-likelihood to our data. 


\vspace{-3mm}
{\small
\begin{flalign}\label{eq:moments}
\begin{aligned} 
     \hat{\alpha} = \hat{\mu}\left[ \frac{\hat{\mu}(1-\hat{\mu})}{\hat{\sigma}^2}  - 1  \right], \,\,
    &\hat{\beta} = (1-\hat{\mu})\left[ \frac{\hat{\mu}(1-\hat{\mu})}{\hat{\sigma}^2} - 1 \right]
\end{aligned}
\end{flalign}
\vspace{-1mm}
}
The parameters, $\hat{\alpha}$ and $\hat{\beta}$, as shown in Eq.~\ref{eq:moments}, are calculated based on the mean $\hat{\mu = \frac{1}{M}\sum_{i=1}^{M} e_i}$ and variance $\hat{\sigma}^2 = \frac{1}{M-1} \sum_{i=1}^{M} (e_i - \hat{\mu})^2$ of the bootstrapped energy values, which determine the shape and scale of the Beta density and specify a plausible range of values of $\theta^{(k)}$,  $\mathcal{L}(\theta^{(k)} | \hat{\alpha}, \hat{\beta}) = \textrm{Beta}(\hat{\alpha}, \hat{\beta})$. This, along with the prior, $p(\theta^{(k)})$, form the acceptance criteria. Detailed explanations to better illustrate the equation are shown in the Appendix~\ref{appendix:proof}. Intuitively, $\mathcal{L}(\theta^{(k)} | \hat{\alpha}, \hat{\beta})$ outputs a likelihood over the domain of $\theta^{(k)}$, given the bootstrapped energy values. Figure~\ref{fig:abc} provides visualization of the behavior of this function at different $\theta^{(k)}$ and energy values. 

\subsection{Sampling the Posterior}
\label{sec:sampling}
Consequently, we can apply an adapted Metropolis-Hastings (M-H) algorithm \cite{turner2012tutorial} in \oursabc to sample from the posterior by replacing the likelihood term with the pseudo-likelihood.
After the execution of algorithm, the output of \oursabc is a set of Monte Carlo samples, $\mathcal{S}_{\theta}^{(k)}$, which approximate posterior distribution, $p(\theta^{(k)} | \mathcal{D}_e, \hat{\pi}^{(k)})$. The full description of the \oursabc algorithm can be found in the Appendix A. 

We execute the \oursabc algorithm with a burn-in period of $B=10$ iterations, and $N=500$ sampling iterations. Additionally, we set $M=5$ for the number of bootstrapped samples at each iteration. We use a $Beta(0.5, 0.5)$ prior, and a $Beta$ proposal distribution with parameters, $\alpha= 4.0$ , and $\beta= 1e-3$, we have explored other Prior distributions and parameters shown in the Appendix~\ref{appendix:prior}.

%% file: experiments.tex
\section{Experiments}\label{sec:exp}

We compare~\oursabc with state-of-the-art OPE and OPR algorithms by evaluating different policies in various gym tasks as introduced in the Appendix~\ref{appendix:experiment}. Below we first briefly introduce experimental settings, including tasks, baselines, and evaluation metrics. 
Then, we conduct ranking and comparison of different candidate policies corresponding to the three tasks introduced in above section:

$\bullet$~\textit{Ranking on average}: We compare \oursabc and other baseline methods to rank policies with clear performance differences in their mean over multiple online rollouts. 

$\bullet$~\textit{Best/worst-case analysis}: To evaluate the ranking performance under worst/best-cases, we further evaluate policies with similar mean performance, calculate their rankings under worst/best-cases performance, and compare \oursabc and other baseline methods.

$\bullet$~\textit{Pairwise comparison}: To validate the capability of \oursabc in differentiating policies with similar online performances and reasoning the comparison with probabilities, we then use our proposed method \oursabc for pairwise comparison between different policies.

\paragraph{More Analysis}\label{sec:sensitive}
Sensitive analysis and case studies are conducted to better understand \ours, including the effect of size and quality of expert data, different similarity measurements, and prior selection choices. Please refer to the Appendix~\ref{app:sec:detailed}. for more investigations. 

\subsection{Experimental Settings}
We first designed our \toy to verify the proposal of estimation the posterior, then, we use~\oursabc and baseline OPE algorithms to solve the OPR problem on widely-used complex environments with discrete or continuous action spaces. Detailed descriptions of the experiment overview and \toy setup are shown in the Appendix~\ref{appendix:toy}.

\paragraph{Baselines and Variants}
We compare~\ours with the first four representative baselines OPE algorithms with their popular implementation ~\footnote{\url{https://github.com/clvoloshin/COBS}}, DICE follows ~\cite{Voloshin2019EmpiricalSO}. Among the methods, Fitted Q-Evaluation (\fqe)~\cite{Le2019BatchPL,Kostrikov2020StatisticalBF} is a Q-estimation-based OPE method that learns a neural network to approximate the Q-function of the target policy by leveraging Bellman expectation backup operator~\cite{sutton2018reinforcement}. \Qpi~\cite{Harutyunyan2016QWO} and Tree-Backup($\lambda$)~\cite{Precup2000EligibilityTF,munos2016safe} can be viewed as two types of generalization from FQE. Model-based method (\MBased)~\cite{paduraru2013off,Kostrikov2020StatisticalBF,fu2021benchmarks} estimates the environment model to derive the expected return of the target policy, and Bayesian Distribution Correction Estimation known as (\dice)~\cite{Yang2020OfflinePS} is the state-of-the-art offline policy ranking method that estimates the posteriors of distribution correction ratios in DICE methods~\cite{Dai2020GenDICEOG,Zhang2020GradientDICERG,Yang2020OffPolicyEV}. It assumes the individuality of policies during the policy ranking process. We use the open-source implementation in~\cite{Yang2020OfflinePS}. 

We also developed a variant of \ours without the probabilistic capacity as \agr, simply measuring the agreement $Agree(\cdot)$ of $\pi$ to the expert $\pi_e$ directly: $p_i = Agree(A_{\pi}, A_{\pi_e}| D_e)$, where $s$ is sampled from the state list $S$ of $D$, and $A_{\pi}$ is a action list generated by $\pi(s)$ in order with the $A_{\pi_e}$ taken by $\pi_{e}(s)$. The $p_i$ represents the performance of policy $\pi_i$, which can be used to rank accordingly in the candidate policy set $\hat{\Pi}$ to get the $O\{\cdot\}$. The experiment uses negative Euclidean Distance for continuous action space and $1-\frac{A_{same}}{A_{total}}$ for discrete, where $A_{same} = \sum_{a_{\pi}^i, a_{\pi_e}^i} \mathds{1}(a_{\pi}^i = a_{\pi_e}^i)$ and $A_{total} = |A|$. Note that the dataset we used fits the common setting in that it does not contain the probability of behavior policy; therefore, the existing IS methods in OPE cannot be utilized as our baselines, and since SOPR requires datasets containing policies with labeled performance, while \ours do not assume knowing this performance beforehand, thus we exclude the method. 

\paragraph{Evaluation Metrics}
We evaluate~\ours and baseline OPE algorithms with two metrics to reflect their accuracy of ranking candidate policies: widely used ranking metric Normalized Discounted Cumulative Gain (NDCG)~\cite{wang2013theoretical}, and Spearman’s Rank Correlation Coefficient (SRCC), adapted by ~\cite{Paine2020HyperparameterSF,jin2022supervised}. Detailed implementation of metrics is introduced in the Appendix~\ref{appendix:metric}. The ranges of NDCG and SRCC are [0, 1] and [-1, 1]  respectively. The higher, the better.

\subsection{Ranking on Average for Policies with Differentiable Mean Performance}\label{sec:experiment1}
We first evaluate \ours and baseline OPE algorithms on multi-level clearly differentiable policies (by mean performance), the policies adopt the same network architecture but are trained with different epochs, we provide a detailed description of the used policies, training steps, validated ground-truth rank in the Appendix~\ref{appendix:trained}, and we have released pre-trained models and training scripts in code. The implementation of the policies is based on a public codebase~\cite{Raffin2021StableBaselines3RR}~\footnote{\url{https://github.com/DLR-RM/stable-baselines3}}. 

\begin{table*}[!ht]
\centering
\small
\caption{Ranking performance on multi-level differentiable policies evaluation w.r.t. NDCG and SRCC. The higher, the better. Mean and standard error across 5 times experiments are shown. \textbf{Best} and \underline{second best} performance are highlighted. \oursabc achieves the top performance.}

\label{tab:trained-model}
\begin{adjustbox}{center, width=0.95\textwidth}
\begin{tabular}{ccccccccc}
\toprule
                              & \multicolumn{2}{c}{\toy} 
                              & \multicolumn{2}{c}{\mcd} 
                              & \multicolumn{2}{c}{\acro} 
                              & \multicolumn{2}{c}{\pend} 
                              \\ 
                              
                              \cmidrule(lr){2-3}
                              \cmidrule(lr){4-5}
                              \cmidrule(lr){6-7}
                              \cmidrule(lr){8-9} 
                              
                              & NDCG        
                              & SRCC         
                              
                              & NDCG          
                              & SRCC          
                              
                              & NDCG            
                              & SRCC          
                              
                              & NDCG       
                              & SRCC        
                              
                              \\ \midrule
\fqe           & \underline{0.8608}$_{\pm\text{0.17}}$          & 
\underline{0.8228}$_{\pm\text{0.45}}$          & 
0.6135$_{\pm\text{0.06}}$         & 
-0.3771$_{\pm\text{0.23}}$          & 0.5970$_{\pm\text{0.33}}$         & 
-0.1000$_{\pm\text{0.85}}$          & 
0.6620$_{\pm\text{0.01}}$          & 
-0.087$_{\pm\text{0.15}}$          
                 \\
\tree          & \textbf{1.0000}$_{\pm\text{0.00}}$ & 
                 \textbf{1.0000}$_{\pm\text{0.00}}$ & 
                 0.7321$_{\pm\text{0.25}}$         & 
                 -0.0629$_{\pm\text{0.60}}$          & 
                 0.6722$_{\pm\text{0.29}}$          & 
                 0.1619$_{\pm\text{0.59}}$          & 
                 0.7562$_{\pm\text{0.23}}$          & 
                 -0.325$_{\pm\text{0.15}}$           
                 \\
\Qpi           & 0.6650$_{\pm\text{0.27}}$          & 
                 -0.274$_{\pm\text{0.64}}$          & 
                 0.7039$_{\pm\text{0.27}}$          & 
                 0.1943$_{\pm\text{0.74}}$          & 
                 0.6108$_{\pm\text{0.32}}$          & 
                 -0.107$_{\pm\text{0.63}}$          & 
                 0.8133$_{\pm\text{0.13}}$          & 
                 -0.039$_{\pm\text{0.11}}$           
                 \\
\MBased        & 0.7004$_{\pm\text{0.21}}$          & 
                 0.0857$_{\pm\text{0.91}}$          & 
                 0.7093$_{\pm\text{0.24}}$        & 
                 -0.0001$_{\pm\text{0.60}}$          & 
                 0.5785$_{\pm\text{0.32}}$          & 
                 -0.164$_{\pm\text{0.73}}$          & 
                 0.5906$_{\pm\text{0.05}}$          & 
                 -0.246$_{\pm\text{0.18}}$           
                 \\ 
\dice	       & 0.5913$_{\pm\text{0.07}}$          &	    
                  -0.466$_{\pm\text{0.31}}$         &	
                  0.9005$_{\pm\text{0.03}}$ &
                  0.2571$_{\pm\text{0.34}}$	 & 
                  \underline{0.9033}$_{\pm\text{0.10}}$ & 
                  0.8571 $_{\pm\text{0.06}}$	 & 
                  0.7251$_{\pm\text{0.19}}$	      & 
                  0.2976$_{\pm\text{0.42}}$	       
                  \\
\midrule
\agr       
               & \textbf{1.0000}$_{\pm\text{0.00}}$ & 
                 \textbf{1.0000}$_{\pm\text{0.00}}$ & 
                 \underline{0.9829}$_{\pm\text{0.01}}$   & 
                 \underline{0.9357}$_{\pm\text{0.02}}$         & 
                 \textbf{1.0000}$_{\pm\text{0.00}}$          & 
                 \underline{0.9950}$_{\pm\text{0.01}}$ & 
                 \underline{0.9421}$_{\pm\text{0.01}}$ & 
                 \underline{0.8190}$_{\pm\text{0.06}}$  
                 \\
\midrule
\oursabc              
                & \textbf{1.0000}$_{\pm\text{0.00}}$ & \textbf{1.0000}$_{\pm\text{0.00}}$ & 
                \textbf{0.9992}$_{\pm\text{0.00}}$         & 
                \textbf{0.9663}$_{\pm\text{0.06}}$          & 
                \textbf{1.0000}$_{\pm\text{0.00}}$         & 
            \textbf{1.0000}$_{\pm\text{0.01}}$          & \textbf{0.9908}$_{\pm\text{0.01}}$          & \textbf{0.9047}$_{\pm\text{0.05}}$          
                \\
                \bottomrule
\end{tabular}
\end{adjustbox}

\end{table*}

Since the policies are from different epochs, they are clearly differentiable candidates for each task. Therefore, we use the order ranked by mean of the accumulated reward of each policy, when deployed for $n=1000$ times running as ground truth $O\{\hat{\Pi}_{mean}\}$. Then we conduct ranking following section of Probabilistic Offline Policy Ranking, where the ranking methods' performances are measured through the evaluation metrics by comparing $O\{\hat{\Pi}\}$ and $O\{\hat{\Pi}_{mean}\}$. Table~\ref{tab:trained-model} contains experimental results in different environments. 

The results in Table~\ref{tab:trained-model} show that \ours achieves a higher rank correlation coefficient and cumulative gain than baseline algorithms, which means \ours can provide ranking results for different policies with higher accuracy. In addition, \ours performs the most stably under all the environments, not having negative rank correlation results in all the tasks, whereas each baseline OPE algorithm has one or more negative rank correlation results. 

\subsection{Best/worse-case Analysis for Policies with Similar Mean Performance}

\begin{figure*}[t!]
    \centering
    \begin{tabular}{cc}
        \includegraphics[width=0.34\linewidth]{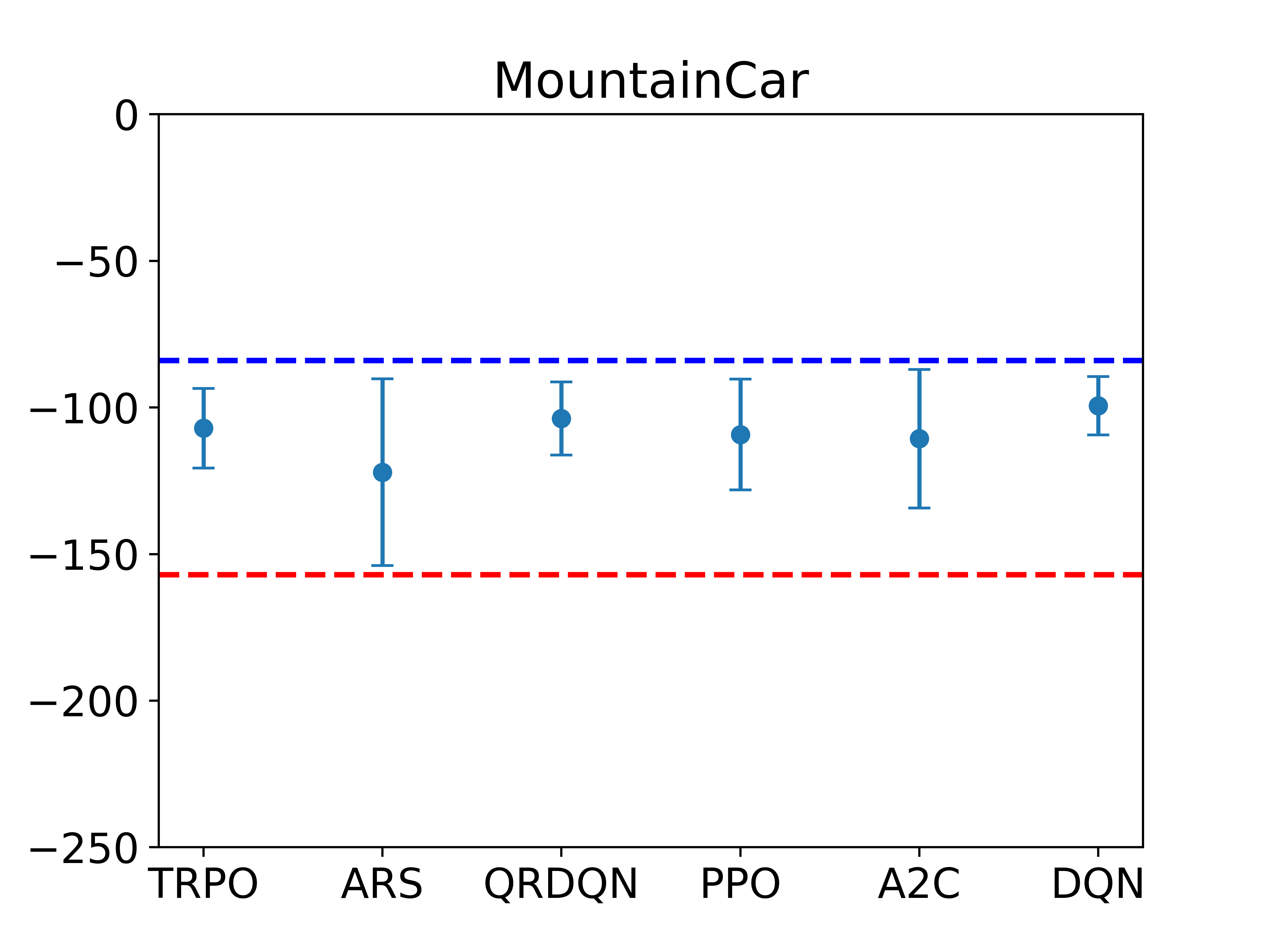}
        & \includegraphics[width=0.61\linewidth]{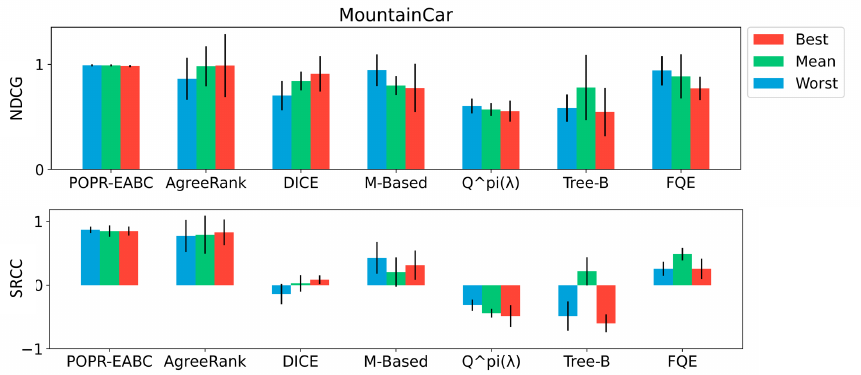} \\
        (a) Online performance. &
        (b) The evaluation results given by different methods.
    \end{tabular}
    \vspace{-2mm}
    \caption{The evaluation on policies with similar mean. (a) The mean and standard deviation of the online performance for different candidate policies. The policies have similar mean, making it hard to rank on the mean. (b) The ranking performance of \oursabc and baseline methods to rank under best/worst/mean case scenario.}
    \label{fig:online:mcd}
    \vspace{-2mm}
\end{figure*}

In this section, we evaluate \ours and baseline OPE algorithms on some open-source policies which show similar mean performance and would like to differentiate the policies through best- or worst-case performance. All the policies are publicly available and well-trained by various RL algorithms, including DQN, QRDQN, TRPO, PPO, A2C, and ARS.~\cite{rl-zoo3}~\footnote{\url{https://github.com/DLR-RM/rl-baselines3-zoo}}. Figure~\ref{fig:online:mcd}(a) shows the mean and standard deviation of their online performance with $10000$ rounds of rollouts in \mcd. It can be found that the performance of different policies is quite similar, while some show larger standard deviations. Under such cases where the candidate policies show similar mean performance, their best/worse case performance would be helpful in practice to decide which is the best. More experimental details and value reports can be found in Appendix~\ref{app:sec:detailed}, and Figure~\ref{fig:opensource-pendulum}.

To get the ground truth ranking of the policies under best/worse case performance, we run each open-sourced policy on the same environment setting for $n=10000$ rounds and log the reward for each round, upon which we take their lowest and highest 5\% values' mean as their worst case and best case performance ground-truth: $O\{\hat{\Pi}_{worst}\}$ and $O\{\hat{\Pi}_{best}\}$. Figure~\ref{fig:online:mcd} shows the results on environments of \mcd, while we also validate in \pend whose results can be found in the Appendix~\ref{app:sec:detailed}. From Figure~\ref{fig:online:mcd}, we have the following observations:

$\bullet$~\oursabc outperforms other baseline evaluation methods with the highest NDCG and SRCC results in all cases, i.e., best, worst, and mean. In the best/worst cases evaluation, \oursabc is able to outperform other baseline methods because it benefits from the performance posterior derived, which we could pay attention to the cases we are caring, while other OPE methods could only produce an expected policy value, and \agr only able to produce an action similarity value. These two groups of approaches, which, under these special cases, fails to effectively and correctly tell the differences.

$\bullet$~\oursabc shows smaller standard deviations under all cases in terms of NDCG and SRCC because it parameterizes the energy value variance in Eq.\ref{eq:moments} and considers such information in pseudo-likelihood to promote a stable and fast convergence to potential posterior.

\subsection{Probabilistic Pairwise Comparison} \label{sec:extremecases}

\begin{figure*}[!ht]
    \centering
    \includegraphics[width=1.0\textwidth]
    {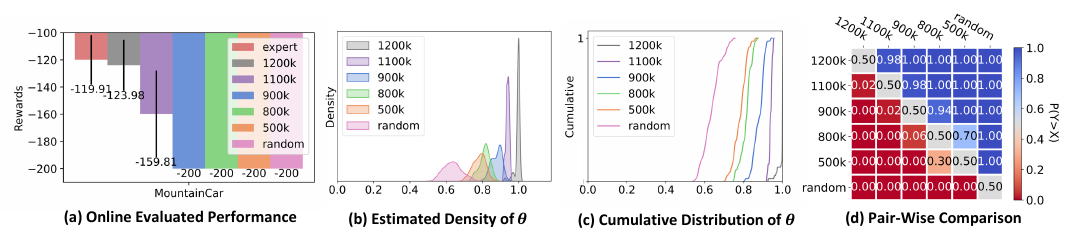}
    \vspace{-7mm}
    \caption{Probabilistic evaluation on policies. (a) The mean performance with standard deviation by rolling out the policies in the online environment. Point estimates are hard to tell the differences between some policies as they have the same mean. (b) Kernel Density Estimates from the posteriors given by \oursabc. (c) The Cumulative Probability Estimates from \oursabc on different policies. The more the line closer to the lower right, the policy performs better. \oursabc can differentiate different levels of policy. (d) The pair-wise comparison on the $\theta$s on different policies. The darker blue the color is, the better the policy from $Y$-axis is than the policy from $X$-axis.}
    \vspace{-4mm}
    \label{fig:probalistic-online-model}
\end{figure*}

Different from existing work, \oursabc can evaluate policies with detailed probabilistic values. Hence, we provide our comparison results from \oursabc on policies trained with different epochs in \mcd as a case study. Figure~\ref{fig:probalistic-online-model}(a) shows the \emph{mean} rewards of each policy using 100 rollouts in the environment. We can see that when the mean rewards are ranked similarly to the policy levels, we sometimes cannot differentiate the performances from the reported identical rewards, because some policies have the same mean in online rollouts shown in (a), while they are at different levels since they are trained with different epochs. This indicates the mean rewards from online rollouts are sometimes inadequate to tell the differences between policies, which requires further evaluations of more than just mean rewards. 

Figure~\ref{fig:probalistic-online-model}(b) and Figure~\ref{fig:probalistic-online-model}(c) present the probabilistic evaluations given by \oursabc with the estimated density of certain PDF, and its cumulative distribution respectively. Since $\theta$ represents the probability of the current candidate policy being as good as the potential expert policy, the faster one reaches 1 (right-top line) in Figure~\ref{fig:probalistic-online-model}(c), the better performance of the evaluated policy, indicating there are more samples of $\theta$ closer to 1 during the evaluation of the policy. 
Reflected by the densities, The approximation for the probability distribution of $\theta$ is in Figure~\ref{fig:probalistic-online-model} (b). Benefit from the (b) characterization of a holistic policy feature, in Figure~\ref{fig:probalistic-online-model} (c), \oursabc can tell the differences between policies using the estimated performance posterior knowledge.

Figure~\ref{fig:probalistic-online-model} (d) presents pairwise comparison of these policies given by \oursabc. Each cell value represents the probability of one policy from the $Y$-axis being better than the other from the $X$-axis. The value of $0$ means $Y$ has probability of zero to be better than $X$ and vice versa when the value is $1$. The results in Figure~\ref{fig:probalistic-online-model} (d) suggest that \ours provides effectively pair-wise probabilistic analysis.


%% file: relatedwork.tex
\section{Related Work}
\textbf{Offline Policy Ranking} is relevant to Q-function selection by choosing the best Q-function from a set of candidate functions. 
Different from OPE, these methods focus on Q-function, whereas in the real world, the target policy may not be in the form of a Q-function. Offline policy ranking has also been studied~\cite{Doroudi2017ImportanceSF,Paine2020HyperparameterSF,jin2022supervised}, which considers point estimates rather than estimating a distribution. Another work~\cite{Yang2020OfflinePS} in OPR that estimates the distribution by transforming it into an oprimization problem with constraints, whereas this paper uses statistical simulation methods to estimate the posterior distribution.
\textbf{Off-policy Evaluation} (OPE) has been focused on estimating the expected value of the target policy. Plenty of OPE methods provide point estimates for the expected value~\cite{Dudk2011DoublyRP,Jiang2015DoublyRO,Zhang2020GradientDICERG,Yang2020OffPolicyEV}. Some OPE methods additionally estimate the value with confidence intervals~\cite{Thomas2015HighConfidenceOE,Kuzborskij2020ConfidentOE,Feng2020AccountableOE,Dai2020GenDICEOG,Kostrikov2020StatisticalBF}. Another direction is estimating and bounding the CDF of returns ~\cite{chandak2021universal, huang2021off}, although these methods are leveraging the distribution estimation, they either require knowledge of action probabilities under the behavior policy or rely on dense returns which restrict the scope of applicability.

%% file: conclusion.tex
\section{Conclusion}
This paper introduces \ours, a framework of a probability-based, statistically-rigorous solution for offline policy ranking. Specifically, we proposed \oursabc to derive the holistic posterior of candidate policies performance as an implementation of \ours, based on an energy pseudolikelihood, it profiles the policy behavior through a probabilistic manner, percept the action variance, and exploits such information in the ABC process, thus, bring awareness to the intrinsic uncertainty of the system. This, in turn, helps estimate the policy's performance efficiently and facilities probabilistic comparisons of candidate policies before they are deployed. We also discuss the potential future work in Appendix~\ref{appendix:limitation}. 

%% file: appendix.tex
\appendix
\section{Full Definition of POPR-EABC Algorithm}

Below we present the full pseudo-code of the POPR-EABC algorithm.

  \begin{algorithm}
    \caption{\oursabc Algorithm}
    \label{algo:abc}
    \DontPrintSemicolon
    \KwIn{Dataset $\mathcal{D}_e$, candidate policy $\hat{\pi}^{(k)}$, prior distribution $p(\cdot)$, proposal distribution $q(\cdot)$, energy function $E(\cdot)$, pseudo-likelihood $\mathcal{L(\cdot)}$}
    \KwOut{Set of posterior samples $\mathcal{S}_{\theta}^{(k)}$}
    
    Initialize posterior $\theta^{(k)}$ and sample set $\mathcal{S}_{\theta}^{(k)}$ \;
    
    \For{$i=1:N$}{
        Get $M$ bootstrapped trajectories $\bar{\mathcal{D}}_e \sim \mathcal{D}_e$ \;
        
        Initialize array of energy values, $\mathcal{E} = \{ \}$ \;
        
        \For{$j=1:M$}{
            Get episode, $ep_j =  \bar{\mathcal{D}}_e[j]$ with length $l$ \;
            
            Initialize synthetic dataset, $\hat{\mathcal{D}} = \{ \}$ \;
            
            \For{$t=1:l$}{
                $s_t = ep_j[t]$, $\hat{a_t} = \hat{\pi}^{(k)}(s_t)$ \;
                
                $\hat{\mathcal{D}}.append([s_t, \hat{a_t}])$
            }
            
            Evaluate energy $e =  E(\bar{\mathcal{D}}_e, \mathcal{\hat{D}})$ \;
            $\mathcal{E}.append(e)$ \;    
            
        }
    
        Calculate $\hat{\mu}$, $\hat{\sigma}$, $\hat{\alpha}$, and $\hat{\beta}$ from $\mathcal{E}$ with Eq.\ref{eq:moments}
    
        Propose new theta $\theta^{\ast} \sim q(\theta^{\ast} | \theta_i^{(k)})$ \;
            
            Compute acceptance probability: \\ $\tau = min \left[1, \frac{\mathcal{L}(\theta^{\ast} | \hat{\alpha}, \hat{\beta})p(\theta^\ast)q(\theta_i^{(k)} | \theta^\ast)}{\mathcal{L}(\theta_i^{(k)} | \hat{\alpha}, \hat{\beta})p(\theta_i^{(k)})q(\theta^\ast | \theta_i^{(k)})} \right]$ \;
            
            \If{$\tau < \phi \sim uniform(0, 1)$}{
                Accept proposal:
                $\mathcal{S}_{\theta}.append(\theta^{\ast})$ \;
                
                $\theta_i^{(k)} \leftarrow \theta^{\ast}$
            }    
            \Else{
            Reject proposal:
            $\mathcal{S}_{\theta}.append(\theta_i)^{(k)}$ \;
            }
        
        Return $\mathcal{S}_{\theta}$ \;
     }
\end{algorithm}

In the \oursabc algorithm, it can be described as a novel ABC algorithm incorporated with Metropolis-Hastings (MH) algorithm. At the beginning, a prior distribution $p(\cdot)$ is defined as $Beta_{prior}$ distribution with parameters as $Beta_{prior}(0.5, 0.5)$, please note that the parameter here is for prior, it is a different set of parameters from proposal distribution $q(\cdot)$ as $Beta_{proposal}$, where $Beta_{proposal}(fix\_{alpha}, epsilon)$, and in our experiment, the $fix\_{alpha} = 4$ and the $epsilon=1e-3$. In the first step, we initialize the posterior $\theta^{(k)}$ by sampling a random number from range [0, 1) using $random.random()$. Then from the expert data set $\mathcal{D}_e$ sample $M$ bootstrapped trajectories as subset $\Bar{\mathcal{D}}_e$. In each subset, traverse the $M$ trajectories, and go through all the given states from the expert dataset, collect the taken behavior from candidate policy, and form the corresponding dataset $\hat{\mathcal{D}}$. Based on the $\hat{\mathcal{D}}_e$ and $\hat{\mathcal{D}}$, calculate the energy value using defined energy function $E(\cdot)$ referring to Section.~\ref{sec:energy_fun}, after the traversal, form a list of energy, $L\_e$, as shown in the Fig.~\ref{fig:intro}, then quantify the mean and variance of the $L\_e$, and feed to the Eq.\ref{eq:moments}, we could obtain the parametrized $\hat{\alpha}$ and $\hat{\beta}$, since the pseudo-likelihood is defined as  $\mathcal{L}(\theta^{(k)} | \hat{\alpha}, \hat{\beta}) = \textrm{Beta}(\hat{\alpha}, \hat{\beta})$, so we could derive term1 $\mathcal{L}(\theta^{(k)} | \hat{\alpha}, \hat{\beta})$ which is a likelihood based on the new proposal $\theta^*$, and together with prior and proposal distribution, and after division following the M-H sampling, the rejecting/accepting phrase will guarantee the algorithm to converge to the desired posterior distribution of $\mathcal{S}_{\theta}$, please note that, each accepted proposal will be used update the value of $\theta_i^{(k)}$, if the proposal is rejected, then $\mathcal{S}_{\theta}$ will add $\theta_i^{(k)}$ rather than the new proposal $\theta^*$.

\section{Summary of Notation}
We first summarize all the important notations used in this paper to help readers to better understand and easier to look up. For representation and meaning please refer to the below Table~\ref{tab:notation}.

\begin{table}[htb]
    \caption{Summary of important notation}
    \label{tab:notation}
    \centering
    \begin{tabular}{c|c}
    \toprule
         Variable & Quantity \\ \midrule
         $i$ & Indexes trajectory  \\ 
         $j$ & Indexes state \\
         $t$ & Indexes time stamp \\
         $k$ & Index of candidate policy \\
         $\hat{\pi}^{(k)}$ & Candidate policy \\
         $A_{n}^{\pi}$ & Action Sequence from policy $\pi$\\
         $A_{n}^{E}$ & Action Sequence from Expert policy logged data\\
         $\pi_e$ & Expert policy \\
         $\mathcal{D}$ & Dataset \\
         $u(\pi)$ & Scoring function\\
         $n$ & Number of trajectories \\
         $L_i$ & Length of trajectory $i$ \\
         $\theta$ & Probability $\hat{\pi} = \pi_e$ \\
         $\tau$ & likelihood of action \\
         $N$ & Number of iterations \\
         $M$ & Number of bootstrap samples \\
         $\mathcal{S}_{\theta}$ & Monte Carlo samples of $p(\theta)$ \\
         $\phi$ & Parameterization Action \\
         $\hat{\mu}$ & Mean \\
         $\hat{\sigma}$ & Variance \\
         

         \bottomrule
         
    \end{tabular}
\end{table}

\section{Discussion on the conclusion at Section~\ref{OPE:conclusion}} \label{sec:reward:proof}

In this section, we will prove that the most of traditional OPE methods align in estimating values, and heavily rely on the dense reward values, which is impractical in practice. This section is to support the conclusion of the main paper in \ref{OPE:conclusion}.

If we define $u(\hat{\pi}^{(k)}|\mathcal{D})$ as a scoring function, that can be used to address the Offline Policy Ranking (OPR) problems, then the OPE could estimate the value to serve for a result of such scoring function.  
Traditional OPE methods aim to estimate the expected value of $V(\hat{\pi}^{(k)}|\mathcal{D})$ precisely with an offline dataset $\mathcal{D}$, which can be described to solve OPR problems by defining the scoring function $\mu(\pi)$ as, e.g., normalized estimated expected value \cite{nachum2019dualdice}, where  $u(\hat{\pi}^{(k)}|\mathcal{D})  \iff   \mathbb{E}[V(\hat{\pi}^{(k)}|\mathcal{D})]$, specifically:
\begin{equation}
 \mathbb{E}[V(\hat{\pi}^{(k)}|\mathcal{D})] = (1-\gamma) \cdot \mathbb{E} [\sum_{t=0}^ \mathcal{H} \gamma^t r_t |s_0, \forall, a_t, r_t, s_{t+1}]
 \label{value_est}
\end{equation}

where initial state $s_0 \sim \beta, \forall, a_t \sim \pi(s_t), r_t \sim R(s_t, a_t), s_{t+1} \sim T(s_t, a_t)$, and $T(s_t, a_t)$ is the transition function. There are multiple methods to derive the value in Eq.\ref{value_est}, like using importance-weighting $\Pi_{t=0}^{\mathcal{H}} \frac{\pi(a_t|s_t)}{\mu(a_t|s_t)}$ \cite{precup2000eligibility}. Thus policy value is calculated by:
\begin{equation}
   \mathbb{E}[V(\hat{\pi}^{(k)}|\mathcal{D})] = (1-\gamma) (\Pi_{t=0}^{\mathcal{H}} \frac{\pi(a_t|s_t)}{\mu(a_t|s_t)} )(\sum_{t=0}^{\mathcal{H}} \gamma^t r_t)
\end{equation}
or leveraging a normalized discounted stationary distribution $d^{\pi}(s,a)$, then calculate $w_{\pi/\mu}(s,a)=\frac{d^{\pi}(s,a)}{d^{\mu}(s,a)}$, and value is calculated by:

\begin{equation}
    \mathbb{E}[V(\hat{\pi}^{(k)}|\mathcal{D})] = \mathbb{E}_{(s,a) \sim d^{\mu}, r \sim R(s,a)}[w_{\pi/\mu}(s,a) \cdot r]
\end{equation}
where the $\pi$ and $\mu$ stand for candidate policy and behavior policy as defined in notations. From the calculation above, we could observe that whichever way is used in OPE, they tend to re-weight or provide correction to the original reward $r$, and finally, estimate the $\mathbb{E}[V(\hat{\pi}^{(k)}|\mathcal{D})]$ of a policy. This will heavily rely on the dense provided reward to guarantee a precise estimation, while in practice, it is often hard to realize.
\section{Explanation on the Pseudo-likelihood}\label{appendix:proof}
To help the reader better understand the Pseudo-likelihood proposed in the paper, we discuss in detail how it parameterizes the variance and mean of the energy values and how the statistical calculation is used in the $Beta$ distribution.

Since the $Beta$ distribution is a flexible and versatile distribution that can accommodate a wide range of shapes and characteristics, making it applicable in many practical situations, we use $Beta$ to capture the features from the energy values derived from the action similarity. And we tend to consider the concentration and variance of the energy values, this intuition drives us to refer to one of the parameterization methods of $Beta(\alpha, \beta)$. If to express $\hat{\alpha}$ and $\hat{\beta}$ in terms of the mean $\hat{\mu}$ and variance $\hat{\sigma}^2$ of energy values $e_i$ in $M$ bootstraps, denote the $\kappa$ as the sum of two parameters:
\begin{equation}
    \kappa = \hat{\alpha} + \hat{\beta} = \frac{\hat{\mu} (1- \hat{\mu})}{\hat{\sigma}^2} - 1
\end{equation}
And the $\hat{\mu}$, $\hat{\sigma}^2$ could be respectively written as:
\begin{equation}
    \hat{\mu} = \frac{1}{M}\sum_{i=1}^{M} e_i
\end{equation}
\begin{equation}
    \hat{\sigma}^2 = \frac{1}{M-1} \sum_{i=1}^{M} (e_i - \hat{\mu})^2
\end{equation}

As defined by $Beta$ that $\hat{\alpha} > 0, \hat{\beta} > 0 $, so we have:

\begin{equation} \label{eq:ineqal}
    \begin{split}
    \kappa = \hat{\alpha} + \hat{\beta} &> 0\\
           \frac{\hat{\mu} (1- \hat{\mu})}{\hat{\sigma}^2} - 1 &> 0 \\
           \hat{\mu} (1- \hat{\mu}) &> \hat{\sigma}^2 \\
           \frac{\hat{\mu} (1- \hat{\mu})}{\hat{\sigma}^2 } &> 1
    \end{split}
\end{equation}

Since there exits the relation that: $\hat{\alpha} = \kappa \hat{\mu}$, $\hat{\beta} = \kappa (1-\hat{\mu})$, if under the condition of Eq.~\ref{eq:ineqal}, we could represent the two parameters as below:
\begin{equation}
    \begin{split}
        \hat{\alpha} &= \kappa \hat{\mu} \\
               &= \hat{\mu} [\frac{\hat{\mu}(1-\hat{\mu})}{\sigma^2}-1]
    \end{split}
\end{equation}
\begin{equation}
    \begin{split}
        \hat{\beta} &= \kappa (1-\hat{\mu}) \\
              &= (1-\hat{\mu})[\frac{\hat{\mu}(1-\hat{\mu})}{\hat{\sigma}^2}-1]
    \end{split}
\end{equation}
We are able to parameterize the $Beta$ distribution based on the above equation into $Beta(\hat{\alpha}, \hat{\beta})$, so the pseudo-likelihood $\mathcal{L}(\theta^{(k)} | \hat{\alpha}, \hat{\beta}) = \textrm{Beta}(\hat{\alpha}, \hat{\beta})$ considers the energy variance, and applying the pseudo-likelihood to the M-H sampling process, \oursabc achieves a high efficiency for the posterior estimation.

\section{Detailed Experimental Settings}
\subsection{Baselines Introduction}
We compare~\ours with four representative baseline OPE algorithms. Four OPE methods with their popular implementation~\cite{Voloshin2019EmpiricalSO}~\footnote{\url{https://github.com/clvoloshin/COBS}}, and DICE as in~\cite{Yang2020OfflinePS}:
\\
\noindent$\bullet$~Fitted Q-Evaluation (\fqe)~\cite{Le2019BatchPL,Kostrikov2020StatisticalBF} is a Q-estimation-based OPE method. It learns a neural network to approximate the Q-function of the target policy by leveraging Bellman expectation backup operator~\cite{sutton2018reinforcement} on a one-step transition iteratively based on the off-policy data.
\\
\noindent$\bullet$~\Qpi~\cite{Harutyunyan2016QWO} can be viewed as a generalization of FQE that looks to the horizon limit to incorporate the long-term value into the backup step.
\\
\noindent$\bullet$~Tree-Backup($\lambda$)~\cite{Precup2000EligibilityTF,munos2016safe} can also be viewed as a generalization of FQE that additionally incorporate varying levels of clipped importance weights adjustment. The $\lambda$-dependent term mitigates instability in the backup step.
\\
\noindent$\bullet$~Model-based estimation (\MBased)~\cite{paduraru2013off,Kostrikov2020StatisticalBF,fu2021benchmarks} estimates the environment model including a state transition distribution and a reward function. The expected return of the target policy is estimated using the returns of Monte-Carlo rollouts in the modeled environment.
\\
\noindent$\bullet$~Bayesian Distribution Correction Estimation known as (\dice)~\cite{Yang2020OfflinePS} is the state-of-the-art offline policy ranking method that estimates the posteriors of distribution correction ratios in DICE methods~\cite{Dai2020GenDICEOG,Zhang2020GradientDICERG,Yang2020OffPolicyEV}. It assumes the individuality of policies during the policy ranking process. We use the open-source implementation in~\cite{Yang2020OfflinePS}. 
\\
\noindent$\bullet$~\agr is a variant of \ours without the probabilistic capacity, simply measuring the agreement $Agree(\cdot)$ of $\pi$ to the expert $\pi_e$ directly: $p_i = Agree(A_{\pi}, A_{\pi_e}| D_e)$, where $s$ is sampled from the state list $S$ of $D$, and $A_{\pi}$ is a action list generated by $\pi(s)$ in order with the $A_{\pi_e}$ taken by $\pi_{e}(s)$. The experiment uses negative Euclidean Distance for continuous action space and $1-\frac{A_{same}}{A_{total}}$ for discrete, where $A_{same} = \sum_{a_{\pi}^i, a_{\pi_e}^i} \mathds{1}(a_{\pi}^i = a_{\pi_e}^i)$ and $A_{total} = |A|$. 

\subsection{Experiment Overview}\label{appendix:experiment}

We introduce an overview of the conducted experiments. Including 5 different gym tasks covering both discrete action space and continuous space. The detailed observation space, action type, and space are shown in the Table below.

\begin{table}[htb]\label{tab:experiment}
\small
\centering
\caption{Experimental tasks investigated in this paper cover various environments with both discrete and continuous action spaces and different state spaces. The numbers indicate the dimensions.}
\begin{tabular}{ccc}
\toprule
Env  & Observation & Action \\ \midrule
\toy &     4       &   Discrete(2)    \\
\mcd &     2       &   Discrete(3)          \\
\acro&     6       &   Discrete(3)            \\
\pend&     3       &   Continuous(1)  \\
\bip &    24       &   Continuous (4)            \\ \bottomrule
\end{tabular}
\end{table}

\subsection{Description on \toy}
\label{appendix:toy}
For simplicity, we first verify the idea of performance posterior estimation on a Toy Environment, here we introduce some details of it. We have the source code in the released files which expose the API for simple implementation and reproduction.

\begin{figure}[!h]
\centering
\includegraphics[width=0.3\textwidth]{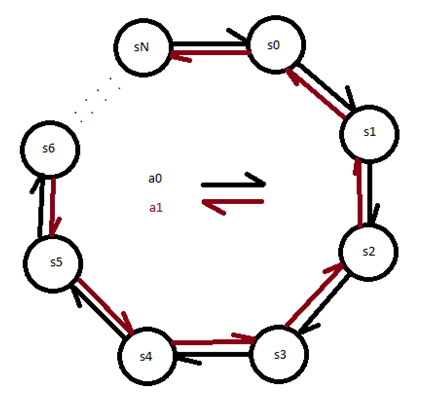}
\caption{Illustration of the N-state toy experiment for preliminary testing}
\label{fig:n-state toy problem graphic}
\end{figure}

We construct a toy environment that consists of a stochastic, two-action space in order to conduct preliminary experiments with POPR. There are n states in a ring. Taking a forward action will move the agent in one direction around the ring while taking the backward action will move the agent in the reverse direction around the ring; however, there is a random chance of any action taken, instead of moving the intended direction, the agent will randomly move to any of the other states (each state has equal probability). For rewards, the agent receives a reward corresponding to the state number so moving to state 0 offers 0 reward while moving to state 1 offers 1 reward, and moving to state n offers n reward. The expert player continuously moves between state n and state n-1 to maximize the reward.

\subsection{Details of Metric Calculation}\label{appendix:metric}
 NDCG comes from DCG, which is a widely used web search metric and is theoretically guaranteed for using the logarithmic reduction factor in Normalized DCG. Assume to an input array with length $n$, we have two possible rankings: $R_x$ and $R_y$, if the items in an order list are not numerical values, we assign natural number values following the rule: highest value $\rightarrow$ list head, and the lowest value $\rightarrow$ list tail. If the ranked value list are $R(X_i), R(Y_i)$, then the SRCC result $r_s$ could be represented as $\rho$,  and we have: 
\begin{equation}
    r_s = \rho_{R(X), R(Y)}= \frac{cov(R(X), R(Y))}{\sigma_{R(X)}\sigma_{R(Y)}}
\end{equation}
where $cov(R(X), R(Y))$ is the covariance of the rank variables. 
In our practical use, since all ranks are attributed to distinct integers, we calculate the formula as:
\begin{equation}
    r_s = 1 - \frac{6\sum d_i^2}{n(n^2 - 1)}
\end{equation}
where $n$ is aligned with the above observation input length, and $d_i = R(X_i) - R(Y_i)$. 
NDCG follows a main DCG formula: $
    DCG_n = \sum\limits_{i = 1}^{n} \frac{2^{rel_i} - 1}{log_2(i+1)}$, 
where $rel_i$ is the relevance score at $i$th element, with a descending relevance as $rel_1=n$ and $rel_n=1$. If we use $R(X)$ to represent the evaluated order and $R(Y)$ to represent the ideal order, we calculate the NDCG as follows:
\begin{equation}
    NDCG = \frac{DCG(R(X))}{DCG(R(Y))}
\end{equation}
$IDCG$ stands for ideal DCG from the empirical ground-truth order list.
The ranges of SRCC and NDCG are [-1, 1] and [0, 1] respectively. The higher, the better.

\subsection{Hyperparameters}\label{app:param}
To guarantee the work's reproducibility, we release all the parameters used in each experiment in this section, besides, we have provided the experiment results in the support files, and for each round of the experiment we have a file named $params.json$ contains the execution automatedly saved param files for reference.

\begin{table*}[htb]
    \caption{Parameters and explanation}
    \label{tab:params1}
    \centering
    \begin{tabular}{c|c|c}
    \toprule
         Parameter name & Explanation & Value \\ \midrule
         $total\_Episodes$ & The amount of trajectory collected for expert dataset & 20\\
         $bootstrap\_M$ & The size of trajectory for every bootstrap execution & 5 \\
         $burnin$ & The amount of discarded sample & 10\\
         $thin$ & The subsampling size after burnin & 10 \\
         $total\_N$ & The total sampling times & 500 \\
         $fix\_alpha$ & The param $\alpha$ if proposal is $Beta$& 4\\ 
         $epsilon$ & The param $\beta$ if proposal is $Beta$ & 1e-3\\
         $Running\_times$ & To execute multiple time for $mean$  and $std$& 5\\
         $expert\_len$ & The default expert trajectory length in Toy Env&  100\\
         $miu$ & The param1 for Norm Prior &  0.4\\
         $sigma$ & The param2 for Norm Prior & 0.4\\ 
         $priori\_type$ & between Norm and Beta &  Beta\\
         \bottomrule
    \end{tabular}
\end{table*}

This table is the parameter table for Section.~\ref{sec:experiment1}. While for the analysis of Section.~\ref{sens:datasize} we keep other parameters, simply change the $total\_Episodes$ as described into 2, 10, 20, 50, 70, 100, and record the results on two metrics NDCG and SRCC. For analysis of Section.~\ref{sens:dataquality}, we mix the default size of 20 with various percentages (vary from 0 to 1, step as 0.1) of expert and non-expert policy trajectory data, and record the evaluation results on the same metrics.
To be clear, for the mixture process in \mcd we use the expert policy as the one trained under 1500k steps and the non-expert policy as the one trained under 500k steps, as shown in Table.~\ref{tab:params1}.

\subsection{Pre-trained Models and Policies}\label{appendix:trained}
Since we conduct experiments on both self-trained policy models and open-sourced policies, so we'd like to introduce the details of our self-trained policies to help understand the process. The table below shows the algorithms used for pre-trained models, followed by used policies and how many timesteps were used for different levels of models, and we evaluated the trained models by running $n=1000$ times online testing and ranked by their performance of normalized accumulated reward for each episode (from start to done), make sure the performance order follows the training time, the longer, the better. However, there is a special case, which is the $MountainCar$ environment, we notice that due to the maximum step restriction for the exploration steps, there are several policies that would perform similarly badly shown as only obtaining -200 reward, we conduct a case study in Section.~\ref{sec:extremecases} of this problem using \oursabc, and demonstrate the effectiveness and benefits of our method. All the pre-trained models are released in the supplementary files together with the code.

\begin{table*}[htb]
\small
    \caption{Pretrained Models and used Policies (1k = 1000)}
    \label{tab:models}
    \centering
    \begin{tabular}{c|c|c|c}
    \toprule
         Environment Name & Used Algorithm & Used Policy & Trained Time Steps \\ \midrule
         $Acrobot$ & PPO &  MLP-Policy & [20k, 40k, 80k, 100k, 200k, 500k]\\
         $BipedalWalker$& SAC & MLP-Policy & [50k, 500k, 1550k, 3150k, 5150k, 6150k, 7750k]\\
         $MountainCar$ & PPO & MLP-Policy & [500k, 800k, 900k, 1100k, 1200k, 1500k]\\
         $Pendulum$ & PPO & MLP-Policy & [100k, 200k, 300k, 400k, 500k, 5000k, 6000k]\\
         \bottomrule
         
    \end{tabular}
\end{table*}

\subsubsection{Open-sourced Policies Details}\label{appendix:opensource}
\begin{figure}[htbp]
    \centering
    \begin{tabular}{cc}
        \includegraphics[width=0.45\linewidth]{pics/opensource/MountainCar.png}
        & \includegraphics[width=0.45\linewidth]{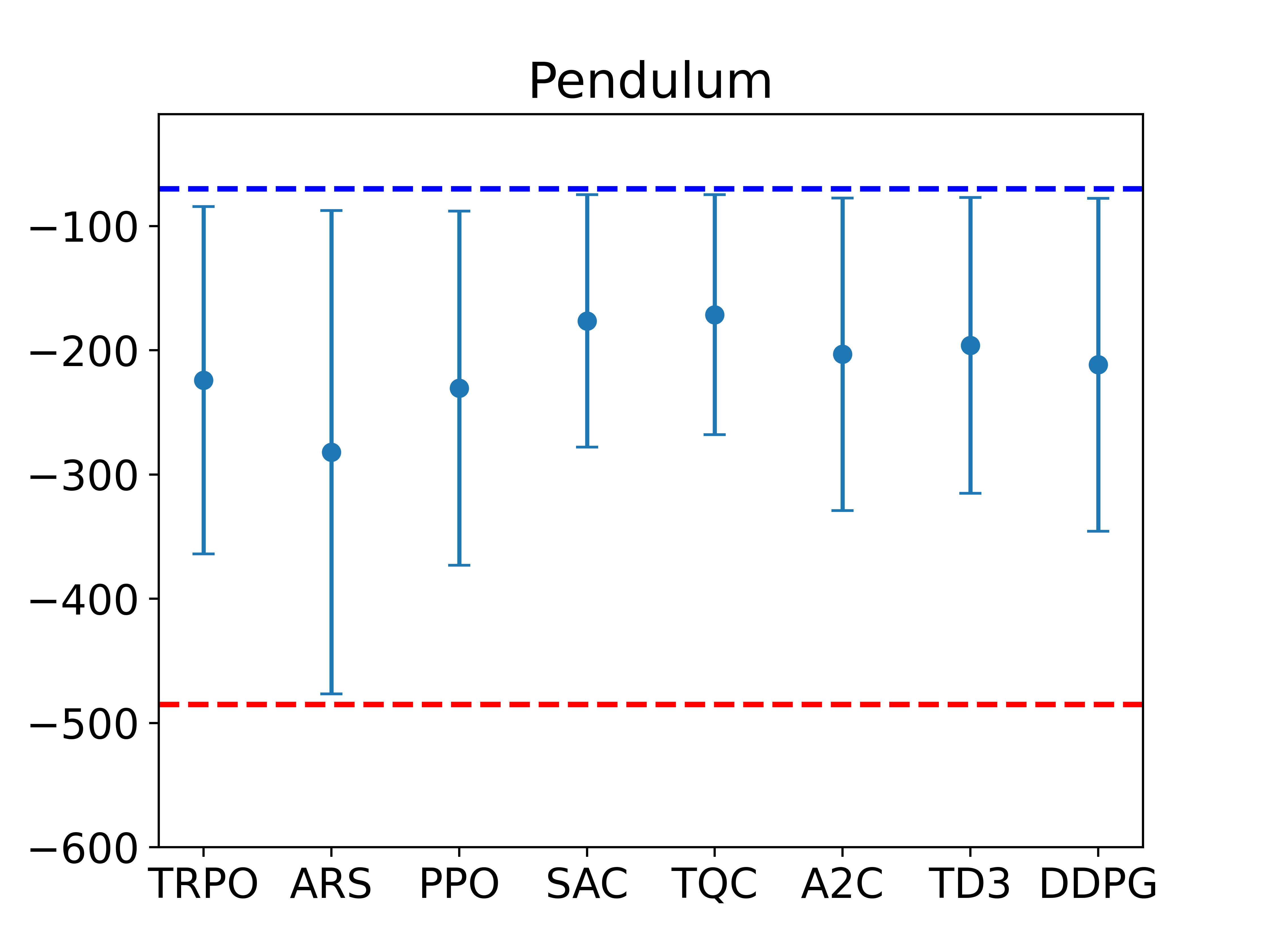}
    \end{tabular}
    \caption{The ground-truth reported by online releases.}
    \label{fig:online}
\end{figure}

In this section, we provide an analysis of open-sourced policies performance and the intuition for the requirement of understanding special cases performance. Then we conduct re-evaluation to get new ground truth $O\{\hat{\Pi}\}_{best}$ and  $O\{\hat{\Pi}\}_{worst}$ under special cases to support the metric calculation for \oursabc and baselines. And we also provide an additional experiment result to evaluate a bag of policies in the \pend environment.
\paragraph{Investigation on the reported performance}
Here we conduct the investigation on the reported values of open source policies, Fig.~\ref{fig:online} shows the online performance in two environments, data are shown in Table.~\ref{tab:reportedmountaincar} and Table.~\ref{tab:reportedpendulum}, where the $y_{axis}$ describes the average reward. We discover that, since all the models were well trained with the very classic RL algorithms, they have a very close mean value, and the variance is even covering the difference of mean, which makes the mean ground truth comparison less meaningful.
This leads us to compare under specific cases, like the worst case and best case study using \oursabc. This is also one of the advantages of our method, benefiting from the posterior estimation.
\begin{table}[htb]
\small
    \caption{Reported Mean Performance of Released Policies for \mcd }
    \label{tab:reportedmountaincar}
    \centering
    \begin{tabular}{c|c}
    \toprule
         Algorithms & Reported Mean and Std \\ \midrule
         $DQN$ & -99.42 $\pm$ 9.92 \\
         $QRDQN$& -103.75$\pm$ 12.50 \\
         $TRPO$ & -107.09$\pm$13.61 \\
         $PPO$ & -109.24$\pm$18.90  \\
         $A2C$ & -110.64$\pm$23.63  \\
         $ARS$ & -122.08$\pm$31.82  \\
         \bottomrule
    \end{tabular}
\end{table}
We also conducted analysis in environment of \pend as in Table.~\ref{tab:reportedpendulum}:\\
\begin{table}[htbp]
\small
    \caption{Reported Mean Performance of Released Policies for \pend}
    \label{tab:reportedpendulum}
    \centering
    \begin{tabular}{c|c}
    \toprule
         Algorithms & Reported Mean and Std \\ \midrule
         $TQC$ & -171.32$\pm$ 96.54 \\
         $SAC$& -176.33$\pm$ 101.55 \\
         $TD3$ & -195.99$\pm$119.03 \\
         $A2C$ & -203.15$\pm$125.77  \\
         $DDPG$ & 211.65$\pm$134.05  \\
         $TRPO$ & -224.04$\pm$139.76  \\
         $PPO$ & -230.42$\pm$142.54  \\
         $ARS$ & -282.084$\pm$194.51  \\
         \bottomrule
    \end{tabular}
\end{table}

\paragraph{Re-evaluation for new ground-truth performance under best/worst cases}
Knowing that we should not only focus on the mean value comparison but also conduct a special case study, and in order to create a surveyor's pole and compare the ranking result from any baseline methods, we need to understand the real performance of these policies to create new groundtruth ranking results. After re-evaluating. In Tabel.~\ref{tab:reportedmountaincar}, we provide the ranking under best and worst cases, we could observe that there exists differences in the ground truth ranking for best cases and worst cases performance, in the environment of  \pend, the changes are more obvious as shown in Table.~\ref{tab:casespndulum}. This indicates that there exists a necessity to consider the extreme cases performances evaluation rather than solely consider the mean value.

\begin{table}[!h]
\tiny
    \caption{Re-evaluate Cases Performance of Released Policies for \mcd }
    \label{tab:evaluatemountaincar}
    \centering
    \begin{tabular}{c|c}
    \toprule
         Case type & Order \\ \midrule
         $Best Case$ & DQN$\geq$ QRDQN $\geq$ TRPO $\geq$ PPO $\geq$ A2C $\geq$ ARS \\
         $Worst Case$& QRDQN$\geq$ DQN $\geq$ TRPO $\geq$ PPO $\geq$ A2C $\geq$ ARS \\
         \bottomrule
    \end{tabular}
\end{table}
\begin{table}[!h]
\tiny
    \caption{Re-evaluate Cases Performance of Released Policies for \pend}
    \label{tab:casespndulum}
    \centering
    \begin{tabular}{c|c}
    \toprule
         Case type & Order \\ \midrule
         $Best Case$ & TD3$\geq$ DDPG$\geq$ TQC$\geq$ SAC$\geq$ TRPO$\geq$ PPO$\geq$ A2C$\geq$ ARS\\
         $Worst Case$& TQC $\geq$ SAC $\geq$ TD3 $\geq$ DDPG $\geq$ PPO $\geq$ TRPO $\geq$ A2C $\geq$ ARS\\
         \bottomrule
    \end{tabular}
\end{table}


Below in Figure.~\ref{fig:opensource-pendulum}, we provide another experiment result in the Pendulum environment, the results show that \oursabc outperforms baseline methods in the worst-, best, mean cases analysis. Indicating a consistent lower $std$ and higher accuracy using \oursabc reflected from the two metrics.

\begin{figure*}[htbp]
    \centering
    \begin{tabular}{cc}
     \includegraphics[width=0.75\textwidth] {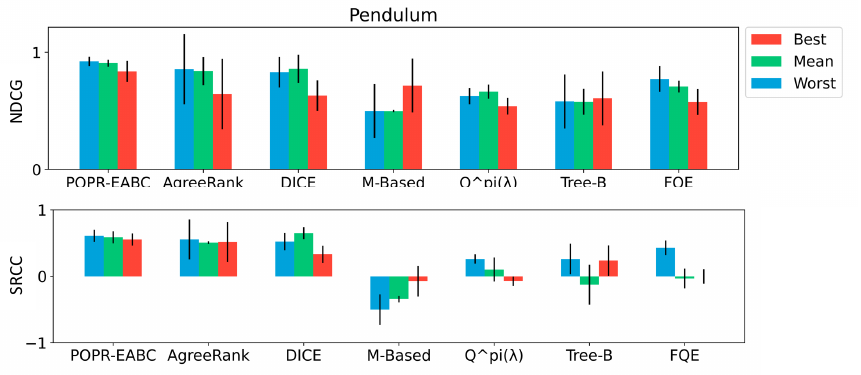}
    \end{tabular}
    \caption{The experiment on open-sourced Pendulum.}
    \label{fig:opensource-pendulum}
\end{figure*}


\section{Detailed Results}
\label{app:sec:detailed}
This appendix section contains very detailed and comprehensive analysis of \ours, specifically, \oursabc.

\subsection{Experiments Results on \toy with Online Learned Policy}
We evaluate \oursabc on multi-level policies, where the policies adopt the same network architecture but are trained with different epochs. 
The ground truth rank labels of these policies are their levels, where a higher-level policy means it is trained with more epochs. From Table~\ref{app:tab:toy-exp} we can see that the mean of $\theta$ can describe the mean performance of different policies.

\begin{table}[htb]
    \caption{Experiment results on \toy for online-trained multi-level policy. Mean and standard error across 10 seeds are shown.}
    \label{app:tab:toy-exp}
    \centering
    \small
\begin{tabular}{cccc}
     \toprule
     \textbf{Policy} & 

     \begin{tabular}[c]{@{}c@{}}$\boldsymbol{\theta}$\\ (\oursabc)\end{tabular}&\textbf{Reward} \\
     \toprule
Expert        & 0.95 (0.04) & 176 (9.6)  \\
2nd Policy    & 0.83 (0.04) & 165 (10.3) \\
3rd Policy    & 0.72 (0.03) & 138 (11.2) \\
4th Policy    & 0.62 (0.05) & 112 (11.2) \\
5th Policy    & 0.53 (0.04) & 114 (11.3) \\
6th Policy    & 0.43 (0.03) & 101 (11.2) \\
Zero Policy   & 0.35 (0.02) & 82 (11.1)  \\
Random Policy & 0.07 (0.02) & 39 (8.8) \\ \bottomrule
\end{tabular}
\end{table}

\subsection{Effects of Data Size}\label{sens:datasize}

\begin{figure}[!h]
    \centering
    \includegraphics[width=0.48\textwidth]{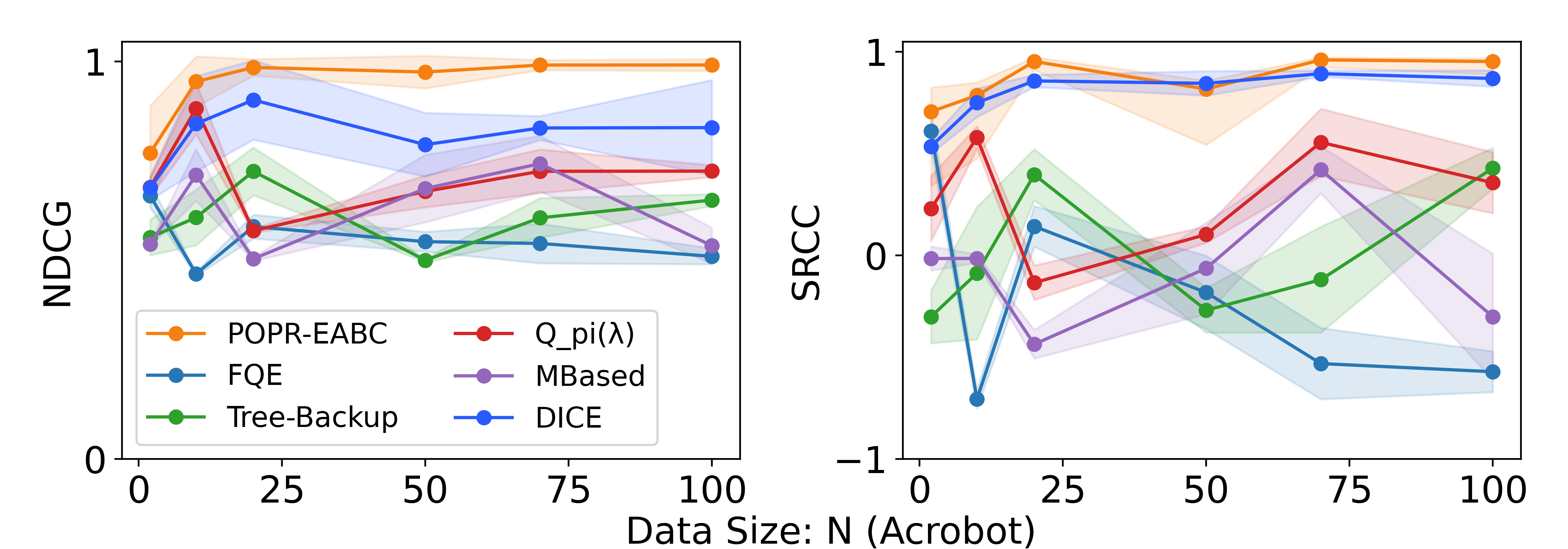}
    \caption{Results in \acro show \oursabc is more robust and effective under different sizes of the dataset.}
    \label{fig:datasize}
\end{figure}

We investigate the sensitivity of \ours to 
the dataset size. We set the number of data trajectories \emph{$\mathcal{D}$} as 2, 10, 20, 50, 70, and 100. The sampling process follows the 
description in Section~\ref{sec:sampling}. We compare with other baseline methods based on the mean and variance shown in Fig.~\ref{fig:datasize}.
We could observe the accuracy of \oursabc is consistently higher than other baselines, and after 20 trajectories, its accuracy reaches stable to almost 1, which justifies our method is robust and effective at small data. It benefits from the efficiency of our method capturing the performance posterior.


\subsection{Effects of Data Quality}\label{sens:dataquality}

\begin{figure}[!h]
    \centering
    \includegraphics[width=0.48\textwidth]{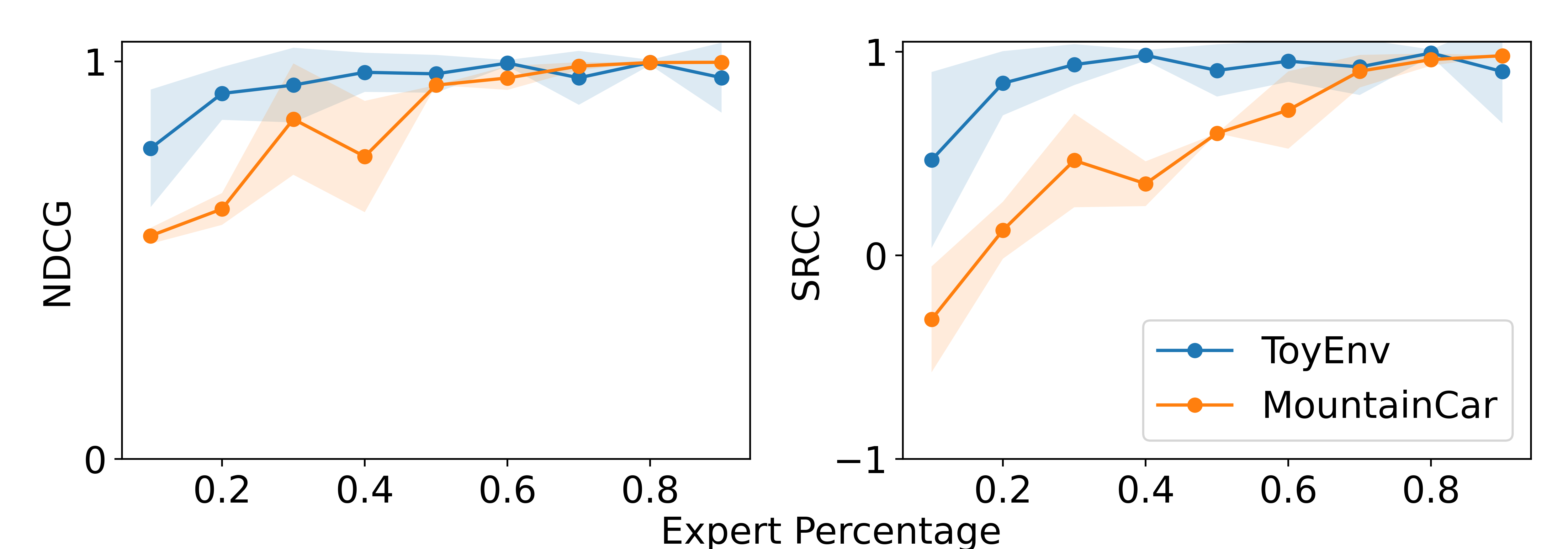}
    \caption{This shows the results over \toy and \mcd of \oursabc with different data quality. \oursabc is effective When the expert takes majority amount of percentage ($\geq$ 0.5).}
    \label{fig:dataquality}
\end{figure}

 We investigate the sensitivity of \oursabc to the quality of the dataset, i.e., whether the dataset is generated by an expert policy or a non-expert policy, and how the percentage of expert policy to all influence the evaluation. In \toy and \mcd, we mix the expert policy data with non-expert ones and vary the percentage of expert data in the dataset from 0 to 1.0. 
 In Figure~\ref{fig:dataquality}, we observe that as the expert percentage increase to above 0.5, the ranking result becomes acceptable, and when above 0.7, results are very promising. This indicates a high tolerance to data quality, which relaxes the expert level of policy and makes it generalizable to real applications. 

\subsection{Effects of Different Choices of $\rho(\mathcal{D}, \hat{\mathcal{D}})$}
\label{app:distance}
In this section, we briefly explore the different choices for $\rho$ measurement and empirically show the effectiveness of Jensen-Shannon divergence, which we finalize as the discrepancy measure utilized in \oursabc.

\begin{table*}[thb]
\tiny
\caption{Effects of Different Choices of $\rho(\mathcal{D}, \hat{\mathcal{D}})$ in \oursabc on the ranking performance with multi-level online-learned policies w.r.t. SRCC and NDCG. The higher, the better.Mean and standard error across 10 seeds are shown. \oursabc with JS divergence achieves the best performance.}

\label{app:tab:online-model}
\begin{adjustbox}{center,max width=\textwidth}
\begin{tabular}{ccccccccccc}
\toprule
& \multicolumn{2}{c}{\toy} & \multicolumn{2}{c}{\mcd} & \multicolumn{2}{c}{\acro} & \multicolumn{2}{c}{\pend} & \multicolumn{2}{c}{\bip}  \\ \cmidrule(lr){2-3}\cmidrule(lr){4-5}\cmidrule(lr){6-7}\cmidrule(lr){8-9} \cmidrule(lr){10-11}
    & NDCG         & SRCC        
    & NDCG         & SRCC         
    & NDCG         & SRCC          
    & NDCG         & SRCC        
    & NDCG         & SRCC       \\ \midrule
                              
JS                   & \textbf{1.0000±0.0} &  \textbf{1.0000±0.0} & \textbf{0.9999±0.01} & \textbf{0.9663±0.06} & \textbf{1.0000±0.0} & \textbf{1.0000±0.01} & \textbf{0.9908±0.01} & \textbf{0.9047±0.05} & \textbf{0.8452±0.14} & \textbf{0.9747±0.03} \\
KL                   & 0.7410±0.24      & -0.062±0.77      
                     & 0.6263±0.26      & -0.050±0.73          
                     & 0.6690±0.32      & -0.021±0.80          
                     & 0.7188±0.22      & -0.007±0.63          
                     & 0.7590±0.16      & 0.3714±0.52     \\
MMD-rbf              & 0.6665±0.28      & -0.125±0.78      
                     & 0.6237±0.11      & 0.0178±0.55          
                     & 0.6322±0.27      & 0.0738±0.49          
                     & 0.5487±0.04      & -0.111±0.31          
                     & 0.6175±0.08      & -0.238±0.59    \\
MMD-multiscale       & 0.8013±0.20      & 0.1942±0.63      
                     & 0.7473±0.23      & 0.0285±0.63          
                     & 0.6908±0.25      & -0.076±0.81          
                     & 0.5250±0.03      & -0.349±0.19          
                     & 0.6701±0.07      & 0.0095±0.31   \\ 
\bottomrule
\end{tabular}
\end{adjustbox}
\end{table*}

Here are different choices of  $\rho$ with the following distance metrics: Jensen–Shannon divergence (JS)~\cite{Endres2003ANM}, Kullback–Leibler divergence (KL)~\cite{Csiszr1975IDivergenceGO}, maximum mean discrepancy (MMD)~\cite{Gretton2012AKT} with an RBF and multiscale kernel. The results in Table~\ref{app:tab:online-model} shows JS divergence works consistently well under different environments. Specifically, JS divergence in this paper uses the base 2 logarithms to make$\rho$ bounded by 1~\cite{Lin1991DivergenceMB}. 

\subsection{Effects of Sampling Times}
\label{app:sample-times}
\begin{figure}[!h]
    \centering
     \includegraphics[width=0.48\textwidth]{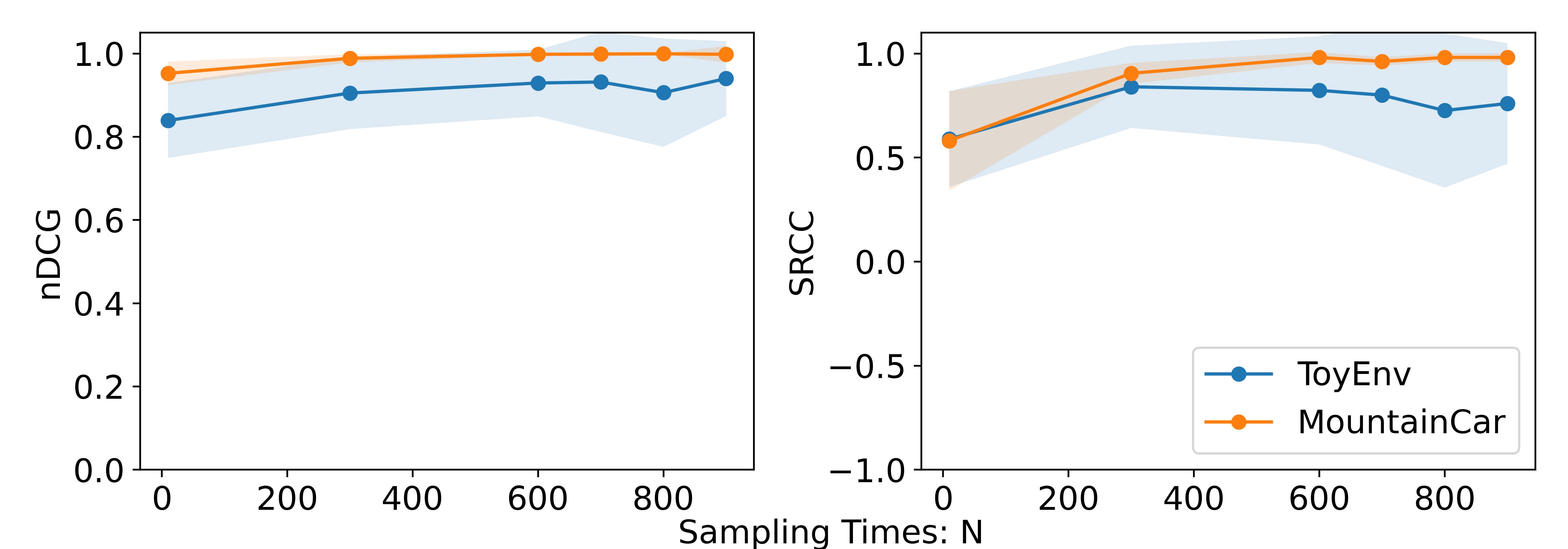}
    \caption{The sensitivity of \ours on sampling times}

    \label{app:fig:sensi-quality}

\end{figure}

 We further investigate the sensitivity of \oursabc to the sampling time \emph{\textbf{N}} as described in Algorithm~\ref{algo:abc}, we tested the sampling times(iterations) in \toy by fixing \emph{$\mathcal{D}$} = 2 and \emph{M} = 1, and in \mcd by fixing \emph{$\mathcal{D}$} = 20 and \emph{M} = 5 and conduct 10 rounds experiment for each setting and present the mean and standard deviation. Note that the evaluation result gap of two lines does not mean \ours performs badly in \toy, the reason causes such gap is the small \emph{$\mathcal{D}$} and \emph{M}, its too quick for \ours to achieve 1 in both metrics if under same setting in \toy as in \mcd, in order to explore the sensitivity, we choose such special design of \emph{$\mathcal{D}$} = 2 and \emph{M} = 1 in \toy. Figure~\ref{app:fig:sensi-quality} shows that increasing sampling time will generally lead to higher accuracy, we found an experiential effective amount on \emph{N} is around 500 times, which is reaching steady evaluation results and prohibiting waste of time. 

\subsection{More Cases on Probabilistic Evaluation}
\begin{figure*}[tb]
    \centering
    \begin{tabular}{cccc}
        \includegraphics[width=0.27\linewidth]{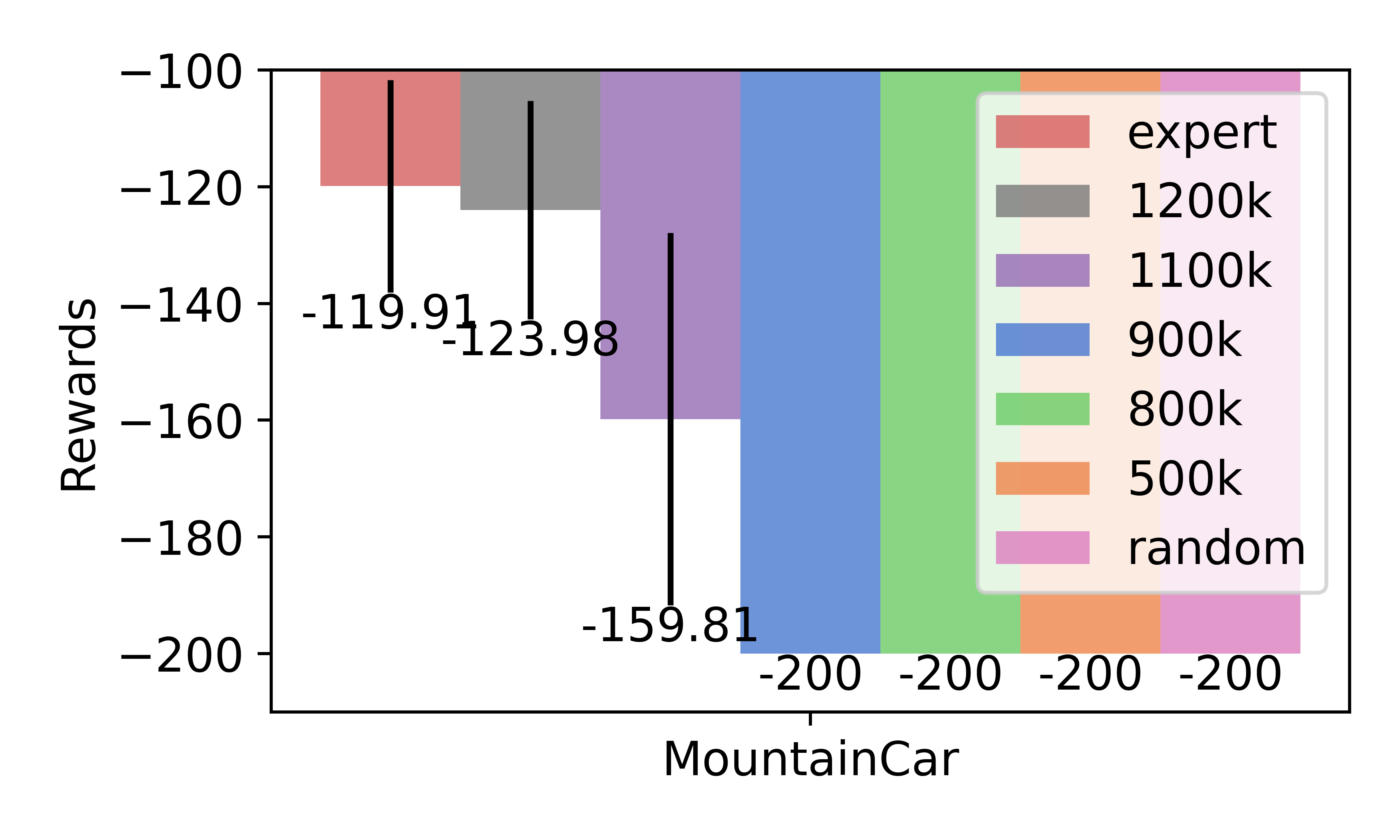}
        & \includegraphics[width=0.22\linewidth]{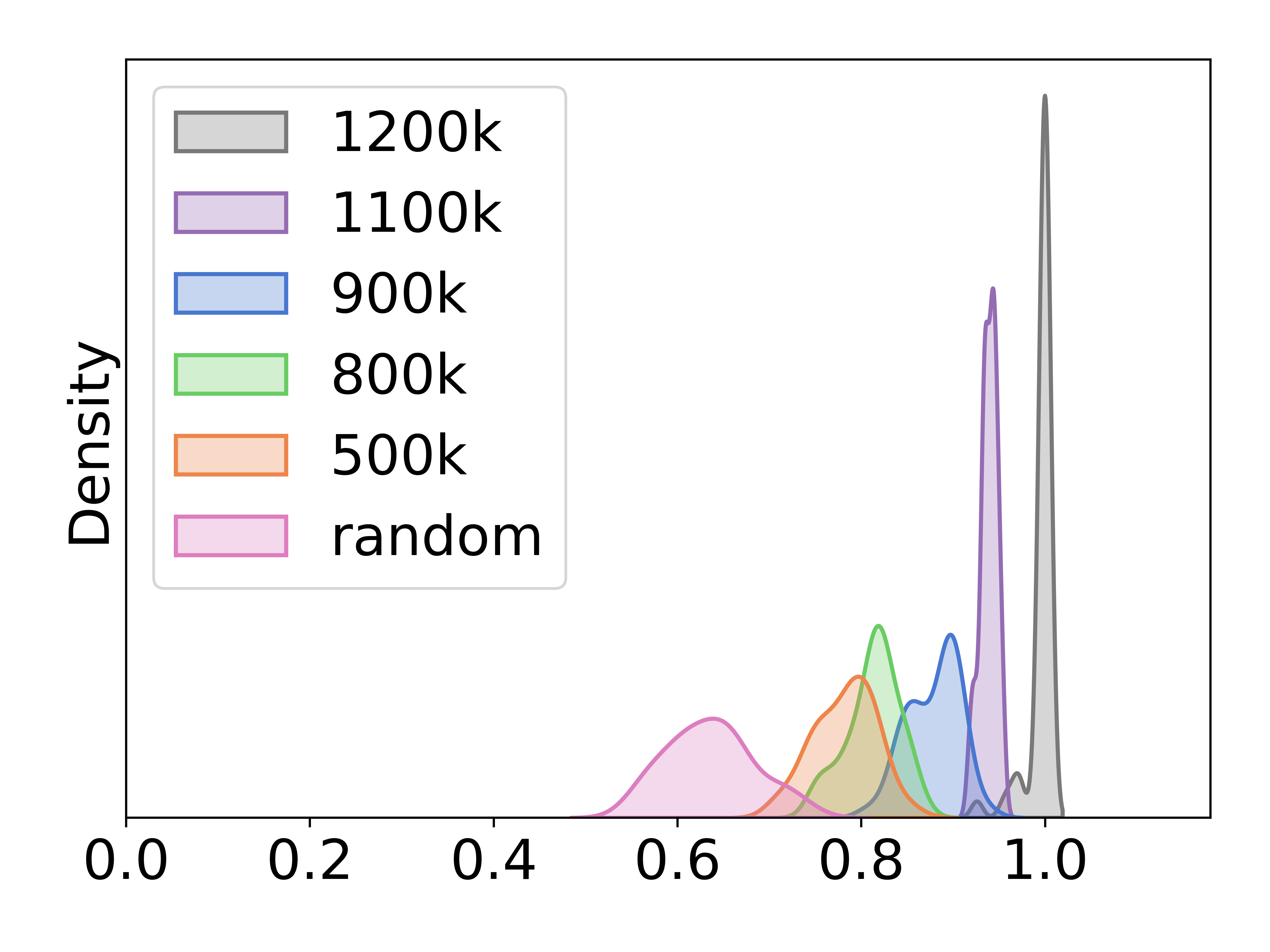}
        & \includegraphics[width=0.22\linewidth]{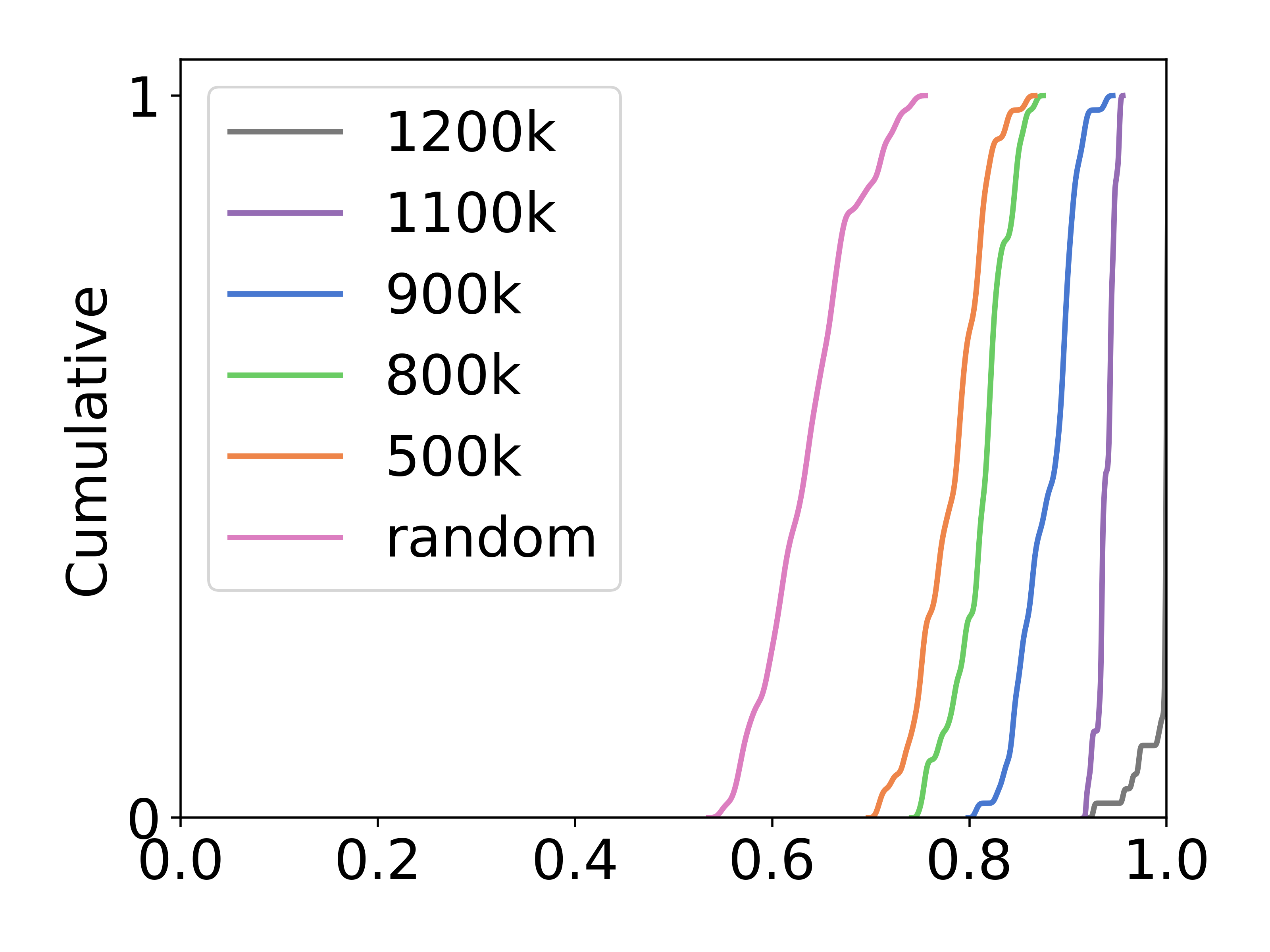}
        & \includegraphics[width=0.22\linewidth]{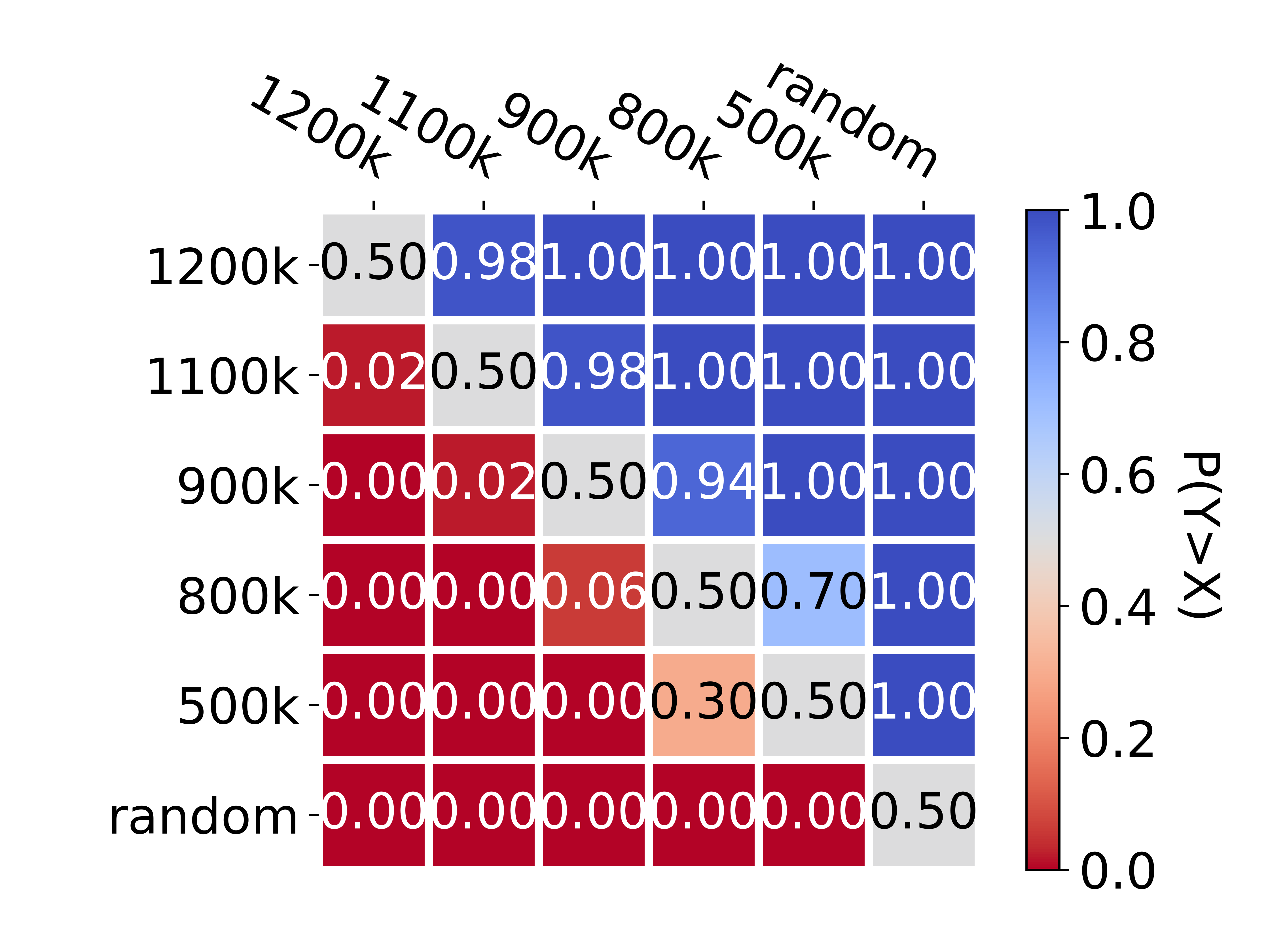}
    \end{tabular} \\
        \begin{tabular}{cccc}
        \includegraphics[width=0.27\linewidth]{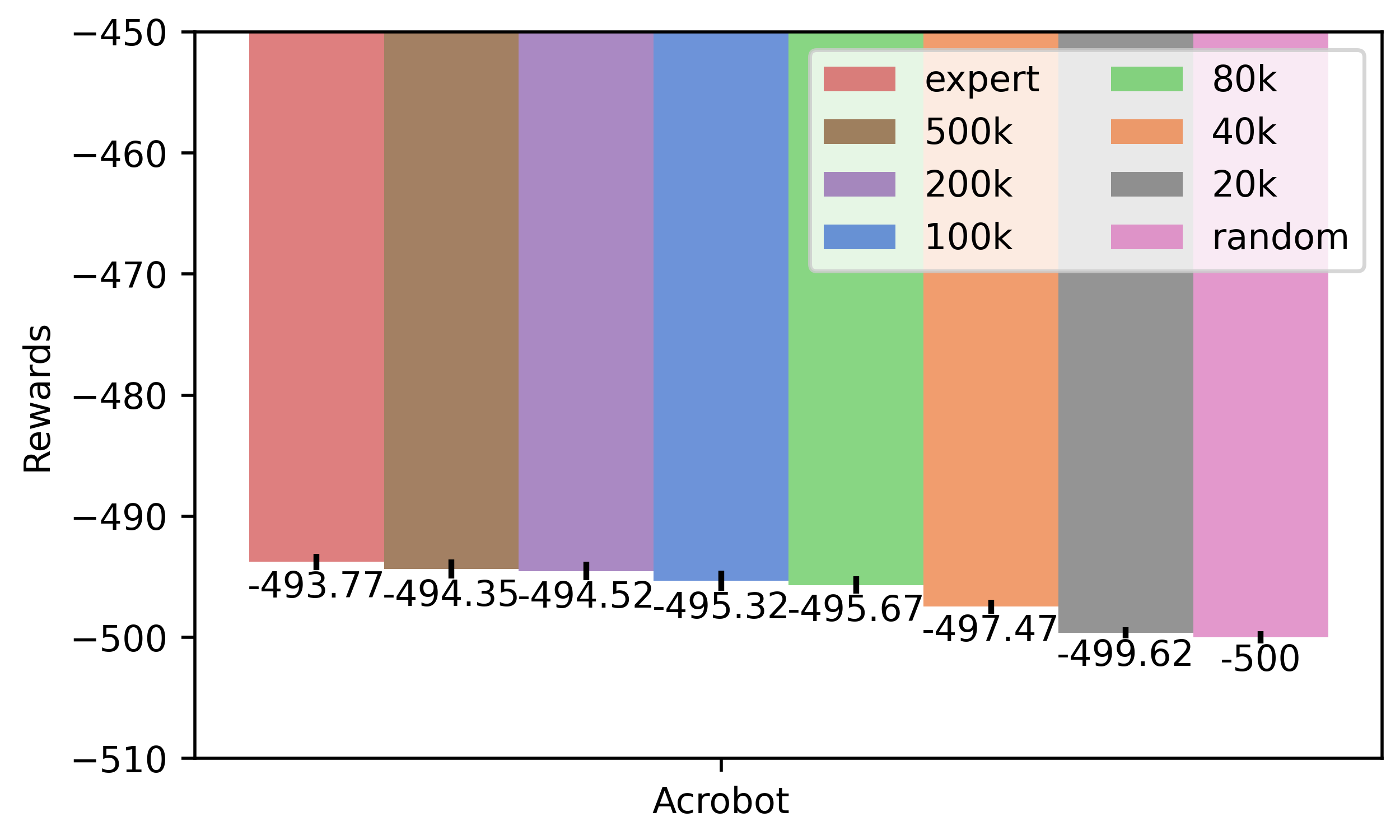}
        & \includegraphics[width=0.22\linewidth]{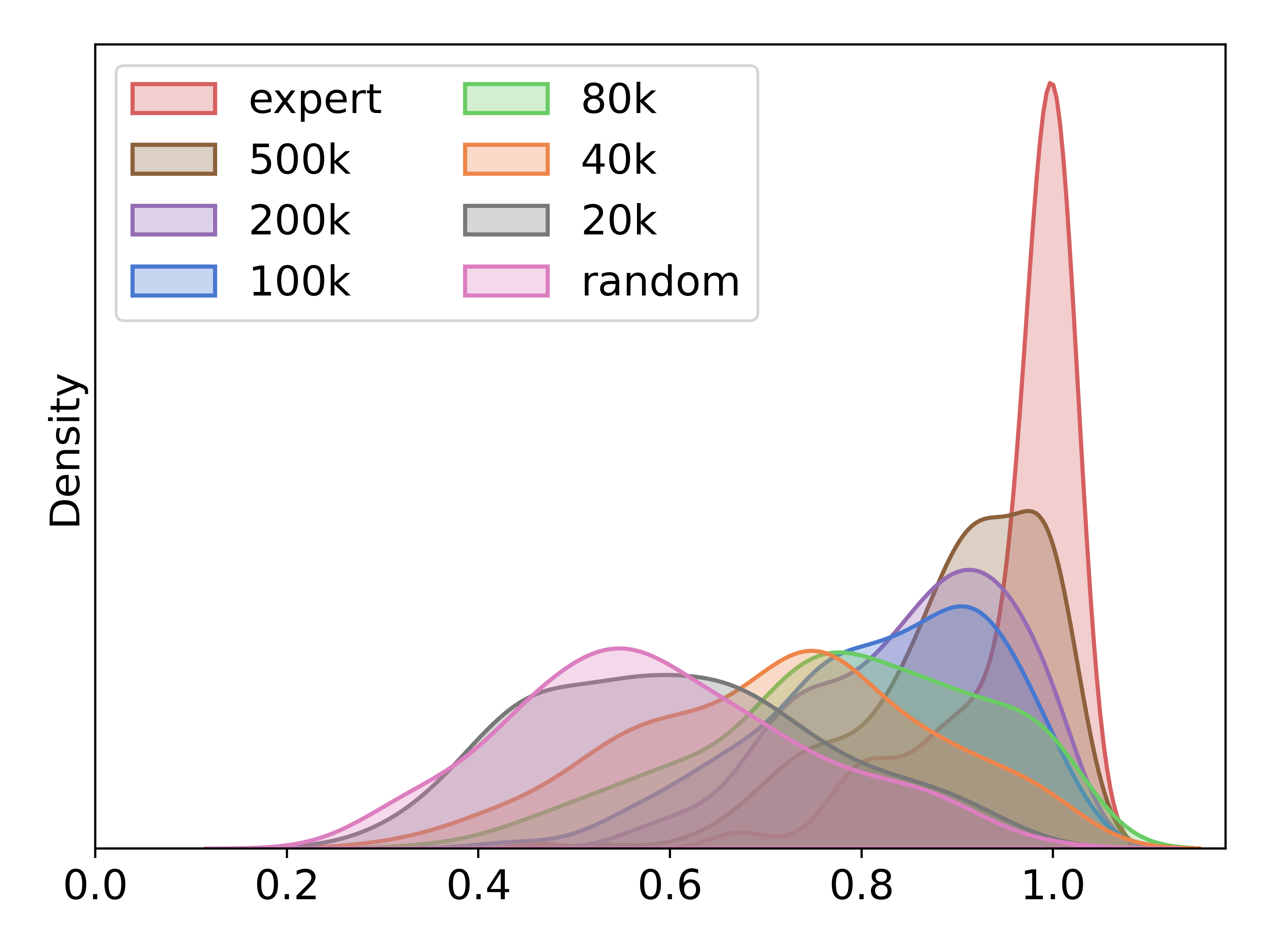}
        & \includegraphics[width=0.22\linewidth]{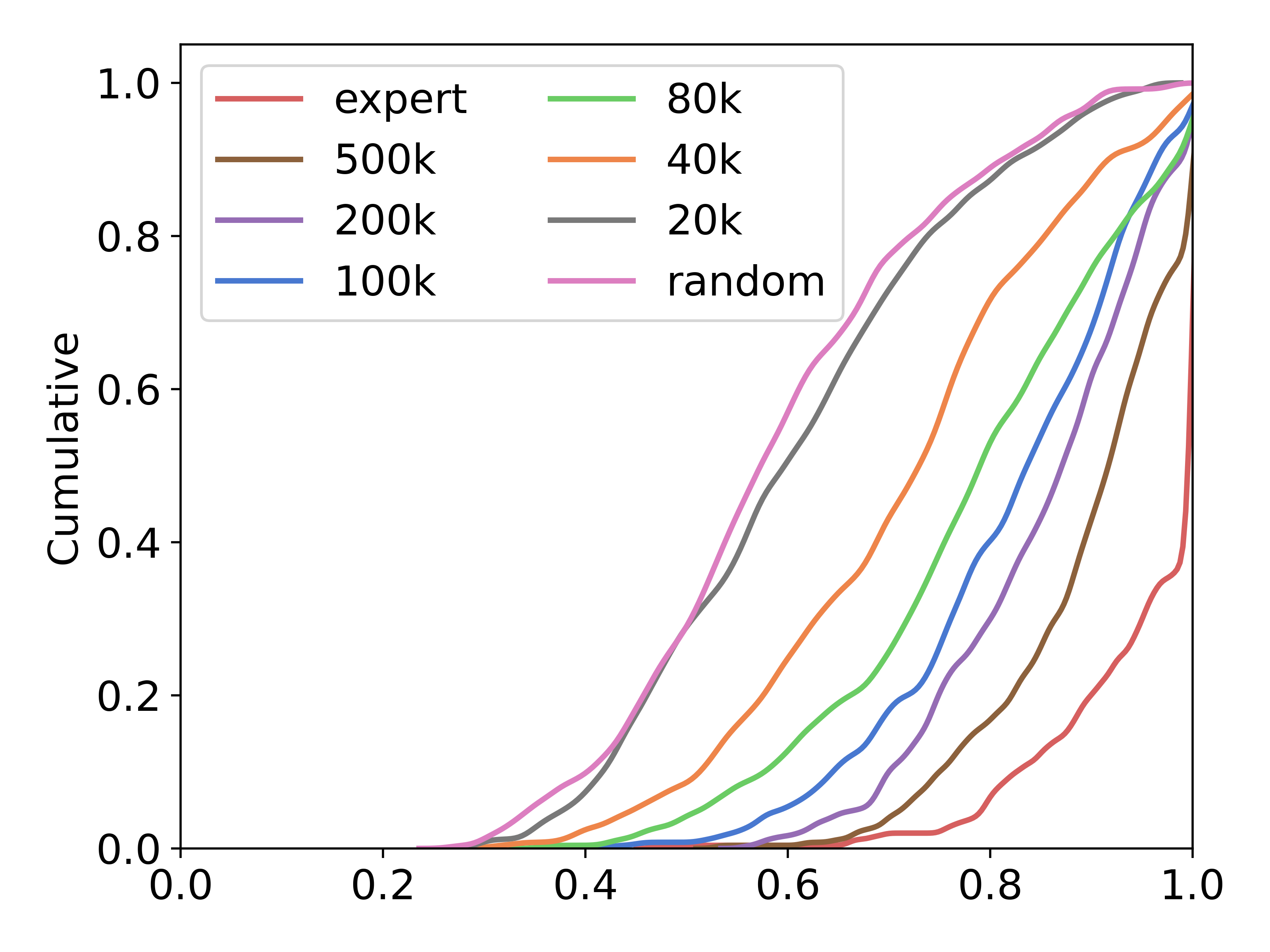}
        & \includegraphics[width=0.22\linewidth]{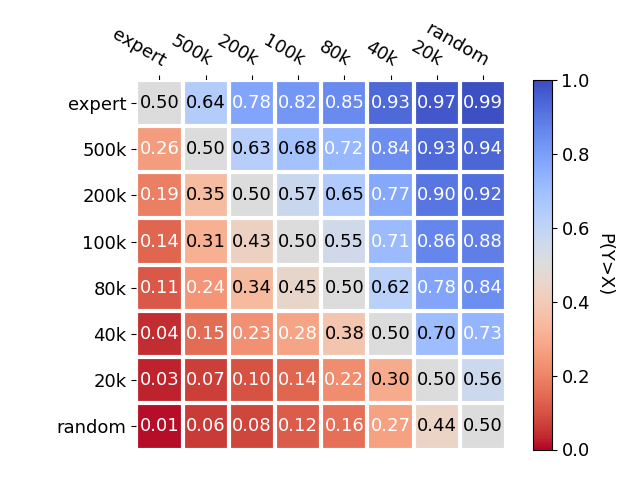}
    \end{tabular} \\
    \begin{tabular}{cccc}
        \vspace{-2mm}
        {\small(a) Online performance} & 
        {\small(b) Estimated density of $\theta$} & (c) Cumulative distribution of $\theta$ &
        {\small(d) Pair-wise }
        \end{tabular} \\
        \vspace{-1mm}
    \caption{Probabilistic evaluation on multi-level online-learned models under two different environments, \mcd and \acro. (a) The mean performance with standard deviation by rolling out the policies in the online environment. (b) Kernel Density Estimates from the posteriors given by \ours. (c) The Cumulative Probability Estimates from \ours on different policies. The more the line closer to the lower right, the better. (d) The pair-wise comparison on the $\theta$s on different policies. The more blue the color is, the better the policy from $Y$-axis is than the policy from $X$-axis.}
    \label{app:fig:probalistic-online-model}
\end{figure*}

Here we provide two cases on probabilistic evaluation under two different environments, \mcd and \acro. Figure~\ref{app:fig:probalistic-online-model}(a) is the mean performance with standard deviation by rolling out the policies in the online environment. Point estimates are hard to tell the differences between some policies as they have the close mean. Figure~\ref{app:fig:probalistic-online-model}(b) is the Kernel Density Estimates from the posteriors given by \ours. Figure~\ref{app:fig:probalistic-online-model}(c) is the Cumulative Probability Estimates from \ours on different policies. The more the line closer to the lower right, the better. \ours can differentiate different levels of policy. Figure~\ref{app:fig:probalistic-online-model}(d) shows the pair-wise comparison on the $\theta$s on different policies. The more blue the color is, the better the policy from $Y$-axis is than the policy from $X$-axis.

\begin{figure*}[thb]
    \centering
    \vspace{-3mm}
    \begin{tabular}{cccc}
        \includegraphics[width=0.27\linewidth]{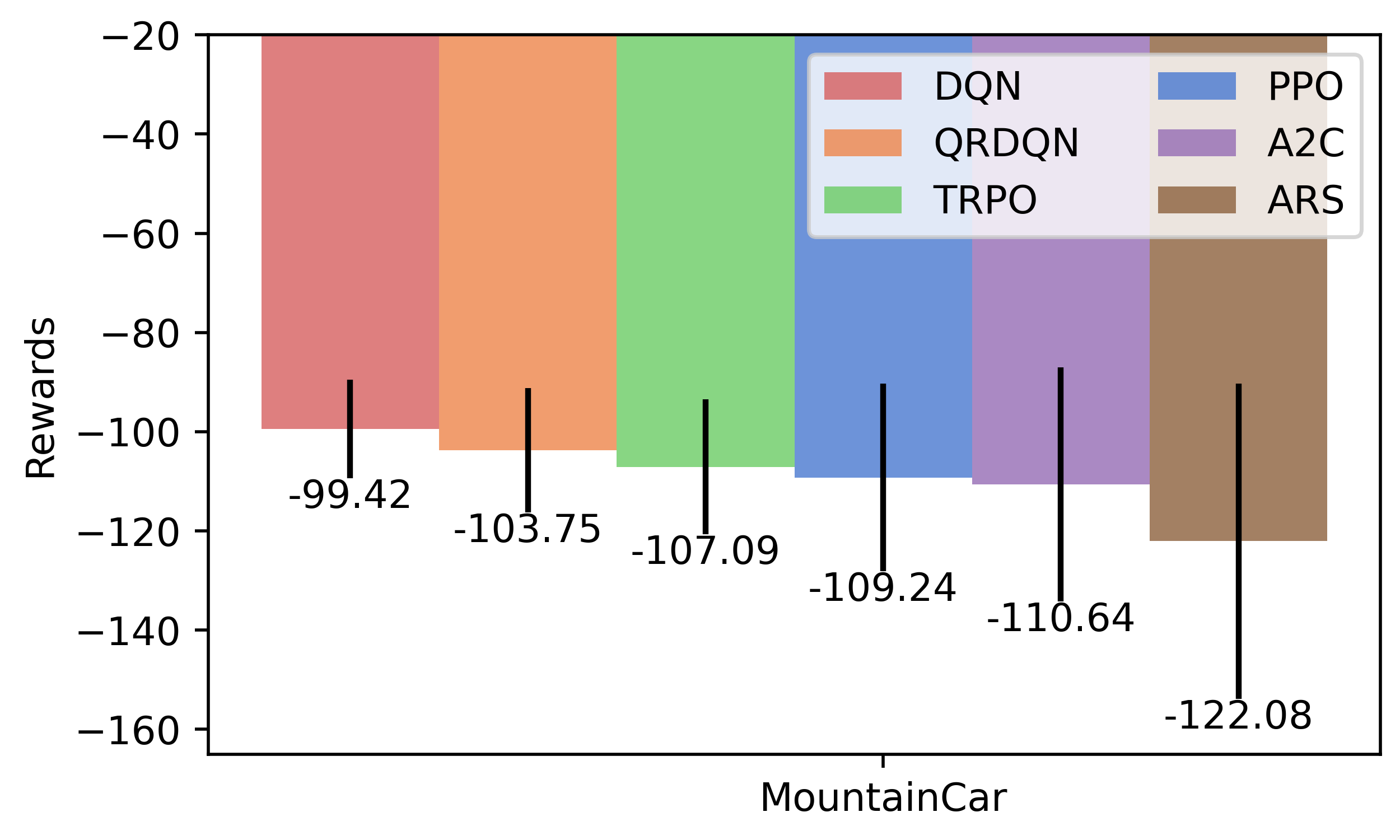}
        & \includegraphics[width=0.22\linewidth]{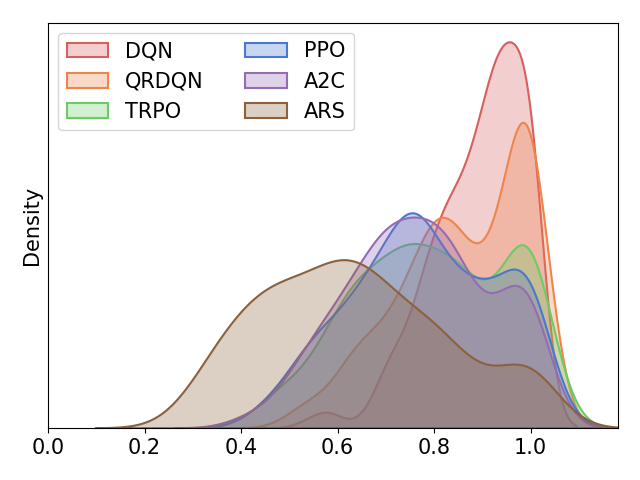}
        & \includegraphics[width=0.22\linewidth]{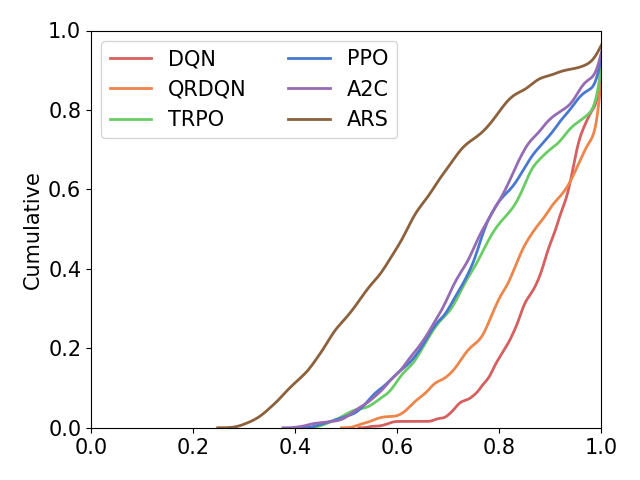}
        & \includegraphics[width=0.22\linewidth]{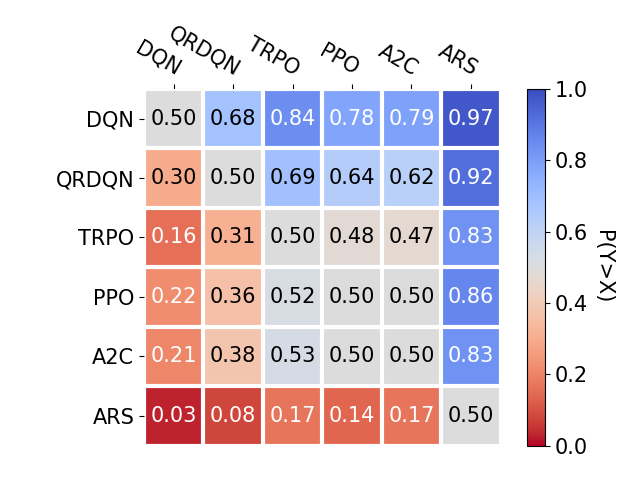}
    \end{tabular} \\
    \begin{tabular}{cccc}
        \includegraphics[width=0.27\linewidth]{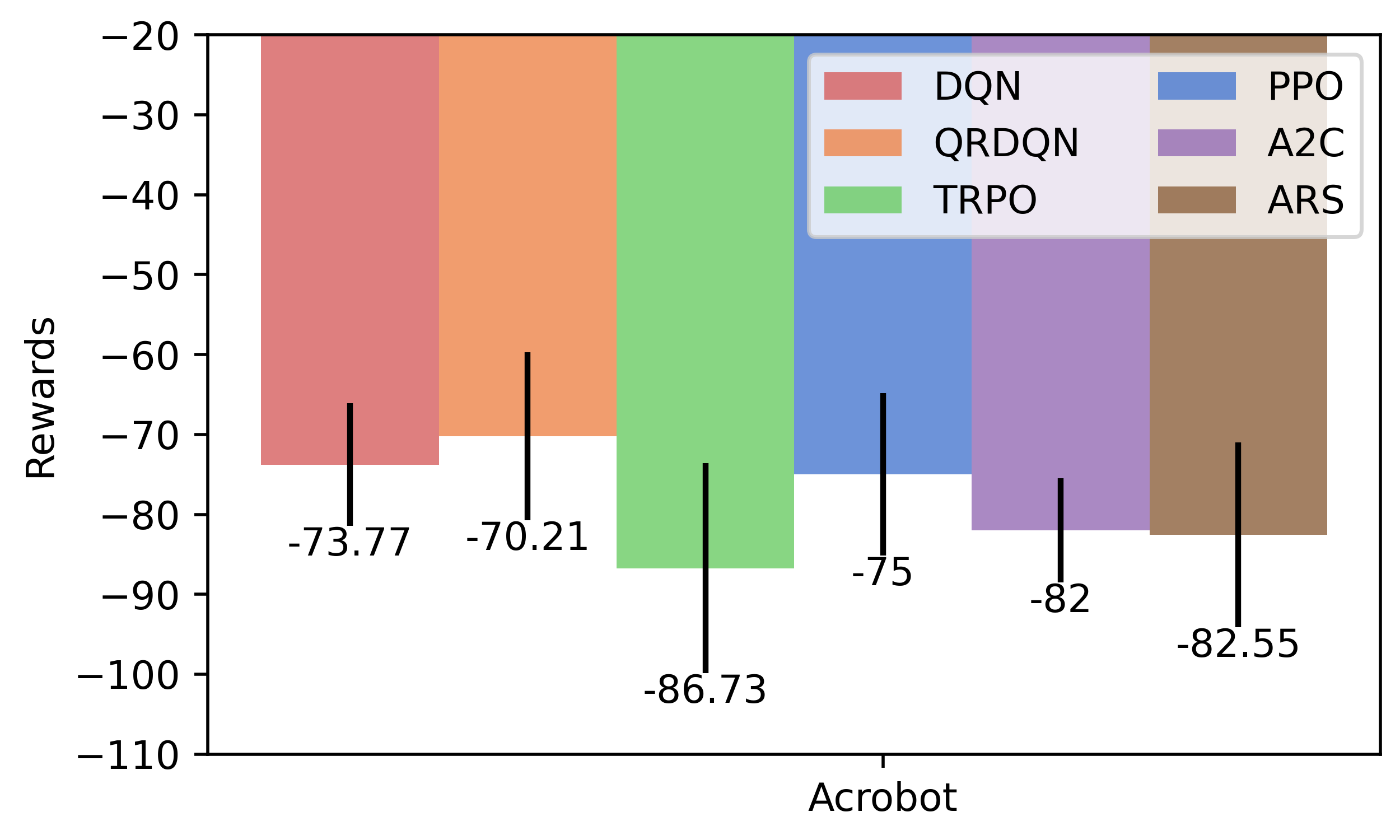}
        & \includegraphics[width=0.22\linewidth]{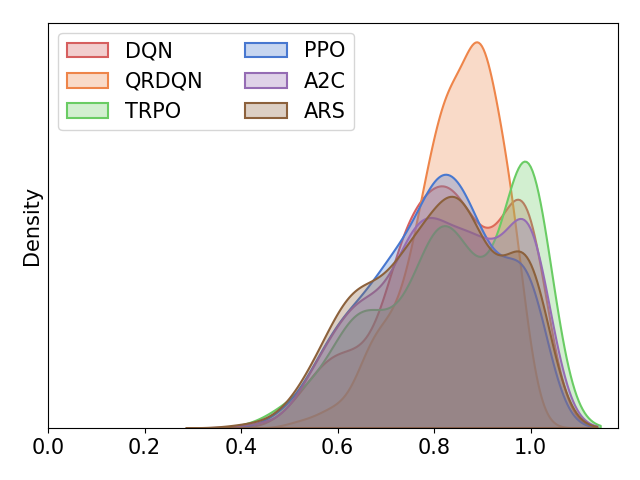}
        & \includegraphics[width=0.22\linewidth]{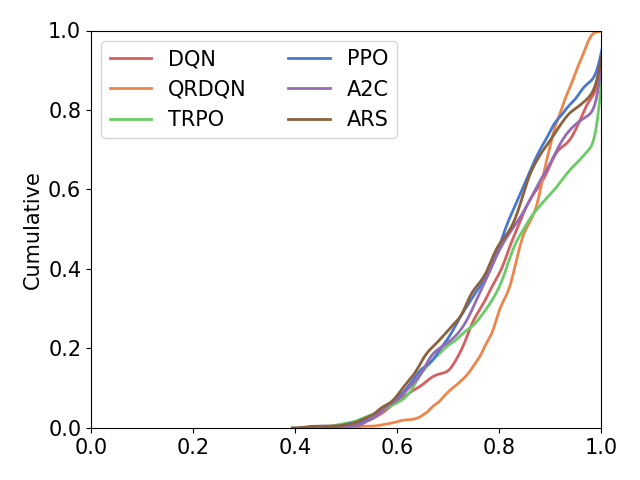}
        & \includegraphics[width=0.22\linewidth]{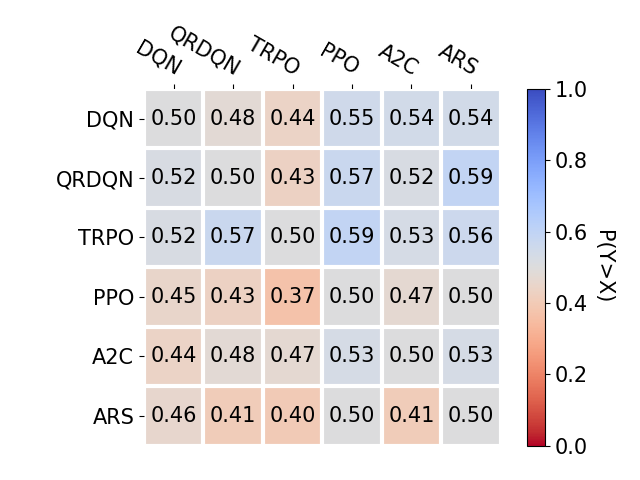}
    \end{tabular} \\
    \begin{tabular}{cccc}
        \vspace{-1mm}
        {\small(a) Online performance} & 
        {\small(b) Estimated density of $\theta$}  & 
        {\small(c) Cumulative distribution of $\theta$}&
        {\small(d) Pair-wise}
        \end{tabular} \\
        \vspace{-2mm}
    \caption{Probabilistic evaluation on open-source policies from StableBaseline3Zoo~\cite{rl-zoo3} under two different environments, \mcd and \acro. Best viewed in color.
    }

    \label{app:fig:probalistic-open-model}

\end{figure*}

\subsection{Extra experiment results on Bipedal Walker}
Due to the page limit, we have appended the BipedalWalker environment policy evaluation results here in the appendix, please refer to Table~\ref{tab:bipdeal}.
\begin{table}[!ht]
\centering
\small
\caption{Rsults of BipdedalWalker}
\label{tab:bipdeal}
\begin{tabular}{ccc}
\toprule Baselines
& \multicolumn{2}{c}{\bip} \\ 

\cmidrule(lr){2-3}
                & NDCG        
                & SRCC     \\ 
\midrule


\fqe  
& 0.7196$_{\pm\text{0.02}}$    
& 0.3452$_{\pm\text{0.26}}$  \\ 

\tree

& 0.6869$_{\pm\text{0.12}}$   
& 0.3177$_{\pm\text{0.25}}$  \\

\Qpi
& 0.7023$_{\pm\text{0.08}}$    
& 0.1975$_{\pm\text{0.22}}$  \\

\MBased
& 0.6128$_{\pm\text{0.11}}$    
& -0.2006$_{\pm\text{0.41}}$  \\

\dice
& 0.8192$_{\pm\text{0.04}}$    
& 0.1437$_{\pm\text{0.18}}$  \\

\midrule
\agr
& 0.8022$_{\pm\text{0.13}}$    
& 0.5417$_{\pm\text{0.15}}$  \\

\midrule
\oursabc
& 0.8990$_{\pm\text{0.05}}$    
& 0.5816$_{\pm\text{0.03}}$  \\

\bottomrule
\end{tabular}
\end{table}








\subsection{Discussion on the Prior and Parameters}
\label{appendix:prior}
In this section, we compare the Norm prior and Beta prior, we also conduct experiments on different parameters for various priors to provide insight on possible solutions for other researchers.

\paragraph{Hyper Parameters for Norm Prior}

\begin{equation}
    f(x) = \frac{1}{\sigma \sqrt{2\pi}} e^{-\frac{1}{2}(\frac{x-\mu}{\sigma})^2}
\end{equation}
Based on the definition of Norm distribution, there are two parameters controlling the shape of the distribution which are $\sigma$ and $\mu$.
We first explored the $\mu$ by randomly fixing the $\sigma = 0.2$, and to observe the best performed parameter value of $\mu$, the purpose of the experiment is to find the most suitable Norm parameters that help the \oursabc to better differentiate the different policies and improve the two metrics results.

In order to understand the performance of using the $Norm$ prior, we conduct the following exploration, note that the initial experiment is conducted on \toy for parameter selections.  
\begin{table*}[!h]
    \centering
        \caption{Exploration on the Parameters of Norm Prior1}
    \label{tab:norm-param}
    \setlength{\tabcolsep}{5mm}{
   \begin{tabular}{cccccc}
    \toprule
    \makebox[0.1\textwidth][c]{Param Group No} & \makebox[0.1\textwidth][c]{$\mu$} & \makebox[0.1\textwidth][c]{$\sigma$} & \makebox[0.1\textwidth][c]{NDCG} & {SRCC} \\
    \midrule
    G1 & 0.1 & 0.2 & 0.863 & 0.771 \\ 
    G2 & 0.2 & 0.2 & 0.750 & 0.314 \\ 
    G3 & 0.3 & 0.2 & 1.000 & 1.000 \\ 
    \textbf{G4} & \textbf{0.4} & \textbf{0.2} & \textbf{1.000} & \textbf{1.000} \\ 
    G5 & 0.5 & 0.2 & 1.000 & 1.000 \\ 
    G6 & 0.6 & 0.2 & 1.000 & 1.000 \\ 
    G7 & 0.7 & 0.2 & 0.996 & 0.828 \\ 
    G8 & 0.8 & 0.2 & 0.996 & 0.828 \\ 
    G9 & 0.9 & 0.2 & 0.759 & 0.371 \\ 
    G10 & 1 & 0.2 & 0.932 & 0.257 \\ 
    G11 & 2 & 0.2 & 0.699 & -0.021 \\ 
    \bottomrule
\end{tabular}}
\end{table*}
\begin{table*}[!h]
    \centering
        \caption{Exploration on the Parameters of Norm Prior2}
    \label{tab:norm-param2}
    \setlength{\tabcolsep}{5mm}{
   \begin{tabular}{cccccc}
    \toprule
    \makebox[0.1\textwidth][c]{Param Group No} & \makebox[0.1\textwidth][c]{$\mu$} & \makebox[0.1\textwidth][c]{$\sigma$} & \makebox[0.1\textwidth][c]{NDCG} & {SRCC} \\
    \midrule
    G4-1 & 0.4 & 0.2 & 0.996 & 0.828 \\ 
    \textbf{G4-2} & \textbf{0.4} & \textbf{0.4} & \textbf{1.000} & \textbf{1.000} \\ 
    G4-3 & 0.4 & 0.6 & 0.937 & 0.428 \\ 
    G4-4 & 0.4 & 0.8 & 0.589 & -0.428 \\ 
    G4-5 & 0.4 & 1.0 & 0.996 & 0.828 \\ 
    G4-5 & 0.4 & 2.0 & 0.994 & 0.942 \\ 
    \bottomrule
\end{tabular}}
\end{table*}

\begin{figure}[htbp]
    \centering
    \begin{tabular}{cc}
        \includegraphics[width=0.45\linewidth]{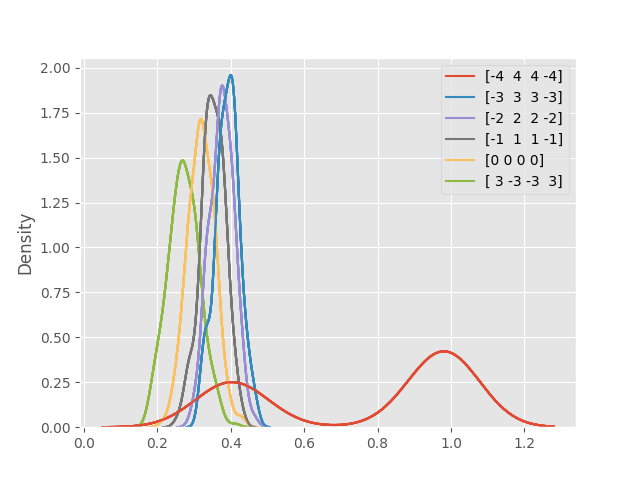}
        & \includegraphics[width=0.45\linewidth]{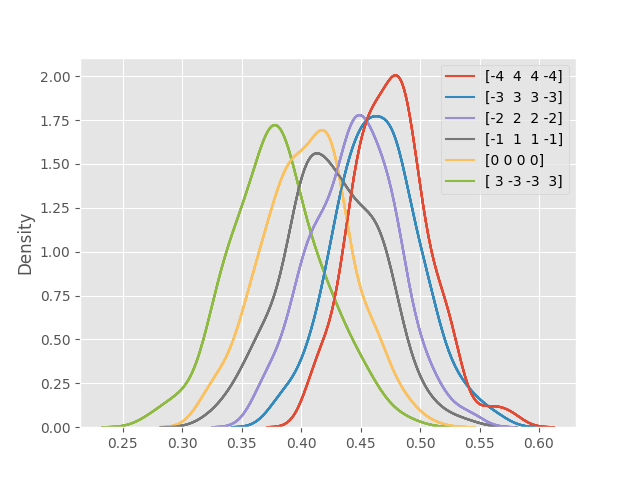}
    \end{tabular}
    \begin{tabular}{cc}
         {\small Norm Prior ($\mu=0.3$, $\sigma=0.2$)}& 
         {\small Norm Prior ($\mu=0.4$, $\sigma=0.2$)}\\
    \end{tabular}
    \begin{tabular}{cc}
        \includegraphics[width=0.45\linewidth]{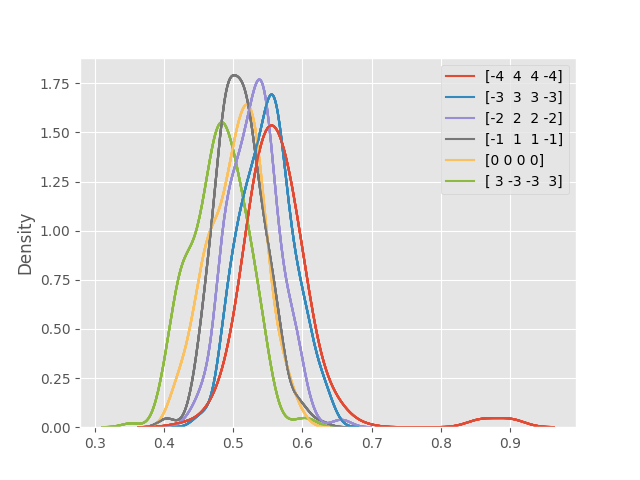}
        & \includegraphics[width=0.45\linewidth]{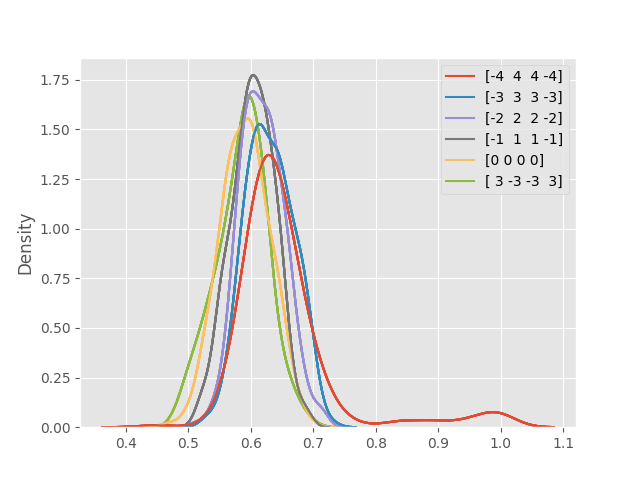}
    \end{tabular}
    \begin{tabular}{cc}
        {\small Norm Prior ($\mu=0.5$, $\sigma=0.2$)} & 
        {\small Norm Prior ($\mu=0.6$, $\sigma=0.2$)} \\
    \end{tabular}
    
    \caption{The parameter exploration on the Norm prior distribution}
    \label{appendix:prior_mu}
\end{figure}

From Table.~\ref{tab:norm-param}, we could observe that there exits G3, G4, G5, and G6 all have a good performance on metrics in \toy, we compared their posterior density figure individually and found that G4 has the most distinguishable shape between policies as shown in Fig.~\ref{appendix:prior_mu} (To avoid the confusion, the labels in the figures are representing the candidate policies' true ability decreasing from red line to green line), so we choose $\mu = 0.4$, similarly we fix the $\mu=0.4$ and explored the $\sigma$ in Table.~\ref{tab:norm-param2}, and come with the conclusion that $\sigma=0.4$ is the most optimal result. So we select the best prior parameter set: Norm($\mu=0.4$, $\sigma=0.4$), and we compare with the current version which used Beta($\alpha=0.5$, $\beta=0.5$) as prior distribution.

\vspace{-4mm}

\begin{table*}[!ht]

\tiny
\caption{Ranking performance comparison using \oursabc with Norm(0.4, 0.4) and Beta(0.5, 0.5) prior respectively, w.r.t. NDCG and SRCC. The higher, the better. Mean and standard error across 5 times experiments are shown. \textbf{Best} performance are highlighted. \oursabc with Beta(0.5, 0.5) prior achieves the top performance.}

\label{tab:priori}
\begin{adjustbox}{center,max width=1.2\textwidth}
\begin{tabular}{ccccccccc}
\toprule
                              & \multicolumn{2}{c}{\toy} 
                              & \multicolumn{2}{c}{\mcd} 
                              & \multicolumn{2}{c}{\acro} 
                              & \multicolumn{2}{c}{\pend} 
                              \\ 
                              \cmidrule(lr){2-3}
                              \cmidrule(lr){4-5}
                              \cmidrule(lr){6-7}
                              \cmidrule(lr){8-9} 
                              
                              & NDCG        
                              & SRCC         
                              
                              & NDCG          
                              & SRCC          
                              
                              & NDCG            
                              & SRCC          
                              
                              & NDCG       
                              & SRCC        
                              
                              \\ \midrule
EABC-Norm(0.4, 0.4)      
               & {0.9438}$_{\pm\text{0.11}}$ & 
                 {0.8228}$_{\pm\text{0.34}}$ & 
                 {0.9809}$_{\pm\text{0.03}}$   & 
                 {0.9583}$_{\pm\text{0.07}}$         & 
                 {0.7984}$_{\pm\text{0.20}}$          & 
                 {0.5396}$_{\pm\text{0.46}}$ & 
                 {0.9468}$_{\pm\text{0.01}}$ & 
                 {0.8603}$_{\pm\text{0.01}}$  
                 \\
\midrule
\textbf{EABC-Beta(0.5, 0.5)}              
                & \textbf{1.0000}$_{\pm\text{0.00}}$ & \textbf{1.0000}$_{\pm\text{0.00}}$ & 
                \textbf{0.9999}$_{\pm\text{0.01}}$         & 
                \textbf{0.9663}$_{\pm\text{0.06}}$          & 
                \textbf{1.0000}$_{\pm\text{0.00}}$         & 
            \textbf{1.0000}$_{\pm\text{0.01}}$          & \textbf{0.9908}$_{\pm\text{0.01}}$          & \textbf{0.9047}$_{\pm\text{0.05}}$          
                \\

                \bottomrule
\end{tabular}
\end{adjustbox}

\end{table*}

\section{Limitations and Future Work }
\label{appendix:limitation}
In this work, we are considering a setting of inaccessible reward value, however, if in any case, the value estimation is important for practitioners, our current proposed approach \oursabc might not be able to directly apply. However, our framework \ours  has the potential of the extension to any value estimation like OPE:
\begin{equation}
    p(R | \mathcal{D}_e, \theta^{(k)}) = \sum_{i=1}^N \sum_{t=1}^{T} p(\theta^{(k)}) r_{it}
\end{equation}
\label{eq:sampled-rew} 

The key idea here is that we can use $\theta$ to calculate an expected reward value.  Once the  posterior distribution of theta, $p(\theta |\mathcal{D}_e)$, has been learned, we can use it to get an estimate of the candidate policy's reward. The Equation in Section.~\ref{eq:sampled-rew} defines the posterior distribution of reward, for non-episodic MDP's \footnote{Episodic MDP's only receive one reward per trajectory, and thus the second summation is suppressed: $p(R | \mathcal{D}_e, \theta^{(k)}) = \sum_{i=1}^N p(\theta^{(k)}) r_{i}$}. Under this formulation, we discount the cumulative expert reward in the dataset, by the probability that candidate policy, $p(\hat{\pi}^{(k)})$, would make the same decision as the expert. We can calculate the posterior mean to get the expected reward under the candidate policy, $\mathbb{E}[R | \mathcal{D}_e, \hat{\pi}^{(k)}]$. By doing so, we should be able to estimate the expected values if desired. We will leave this for future work.

Another limitation could be the assumption of access to the expert data, but the expert policy could be non-unique, which means that the expert policy could be multiple, so this would require a consideration of the coverage of diversity of expert action, we would leave this for future work to discuss.  For the possible question on the quality of expert data (or the expert level of demonstration data), in this paper, as shown in Section.~\ref{sens:datasize}  and Fig.~\ref{fig:datasize}, we have explored the requirement for expert data quality by mixing the noisy policies, and show that it will be beneficial using our methods as long as the majority coverage is expert data.

\section{Preliminary Exploration on Extension to Multiple Experts Policies}

From our analysis of current \oursabc, it only leverages the single expert policy while sometimes, we are able to make the most of multiple expert policies (direct reflection as trajectories). To adapt to this application scenario, we provide an initial study on how to extend our current work to multiple expert policy solutions.

We first verify the data we collect follows the multi-expert policy distribution (existing diversity and variance), and then we propose an architecture that could address the multiple expert policies extension, then we implement an initial case and show the results of implementation.

\paragraph{Verify of Multi-expert Policy Distribution}

We adopted 5-25 well-performed expert policies in MountainCar and Acrobot Environment, below is an example on 5 policies on MountainCar, which guarantees their variety of policies reflected by the various action preference on the same states.

\begin{figure}[!ht]
    \centering
    \includegraphics[width=0.42\textwidth]{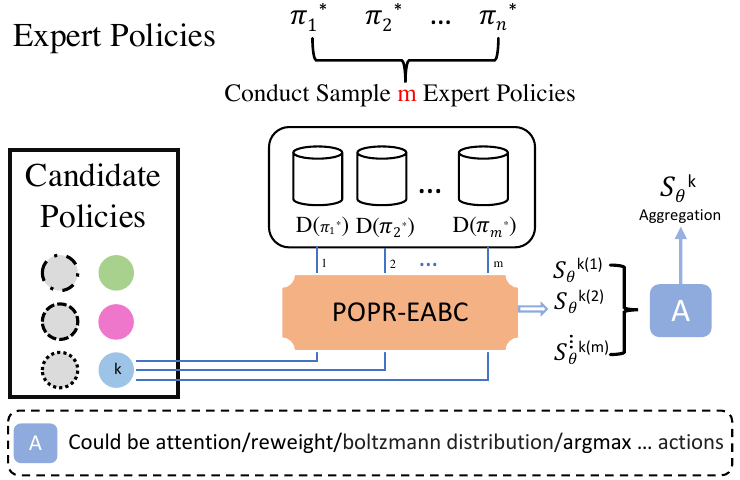}
    \caption{Possible Extension Based on POPR-EABC}
    \label{fig:extension}
\end{figure}

\paragraph{Extension Based on Effective \oursabc}
In this part, we provide a demonstration of how to solve multiple expert policies situation as shown in Fig.~\ref{fig:extension}.
\vspace{-2mm}
\begin{figure}[!ht]
    \centering
    \includegraphics[width=0.40\textwidth]{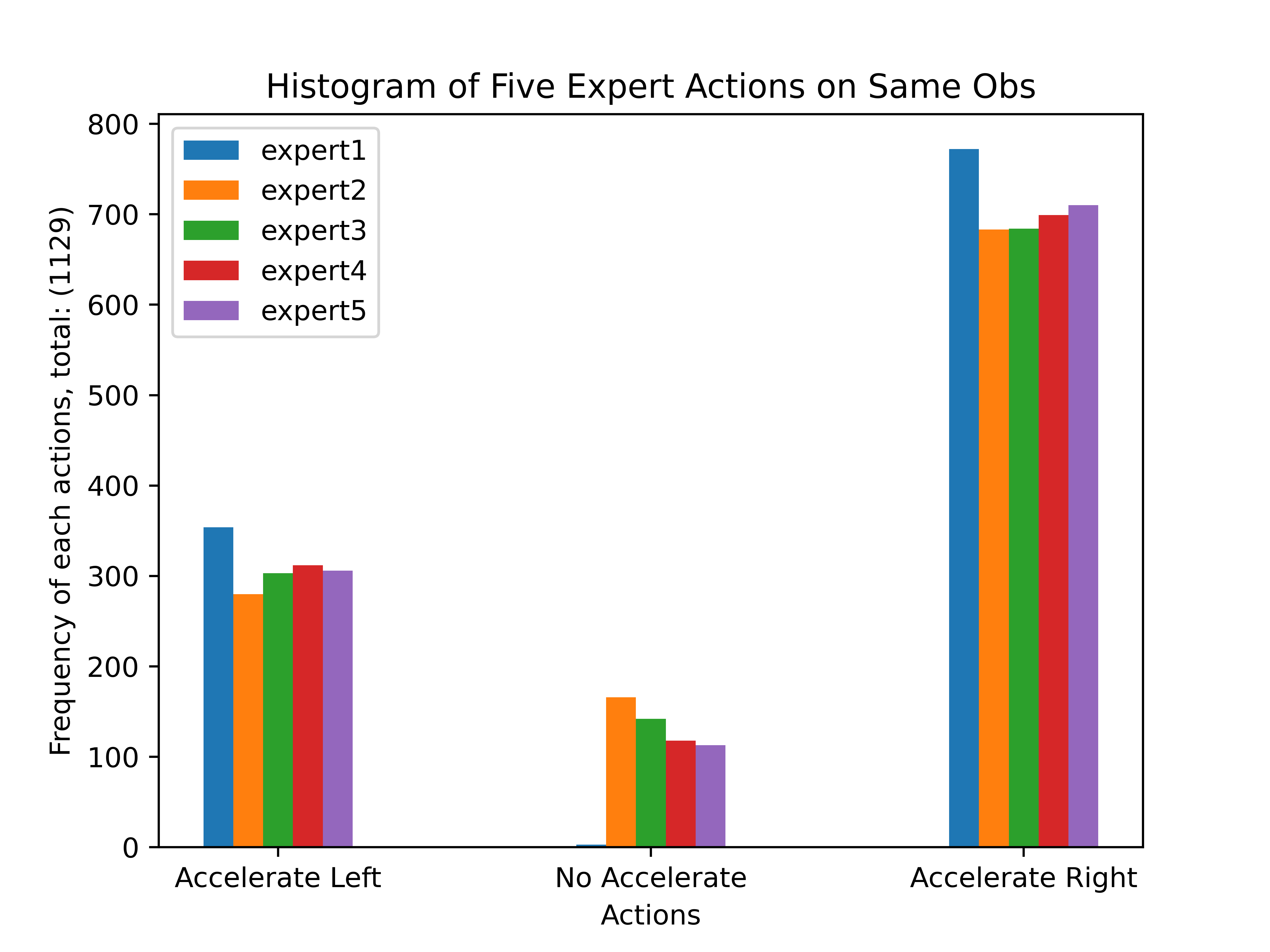}
    \caption{Verification on the expert dataset with policy variance}
    \label{fig:variance}
\end{figure}

\begin{table}[thb]
\tiny
\centering
\caption{Rsults of Multiple Expert Extension on Policies Datasets
}
\label{tab:extension}
\begin{tabular}{ccccc}
\toprule Dataset $D_e$
& \multicolumn{2}{c}{\mcd} & \multicolumn{2}{c}{\acro} \\ 

\cmidrule(lr){2-3}\cmidrule(lr){4-5}
                & NDCG        
                & SRCC        
                & NDCG          
                & SRCC   
\\ 
\midrule

$D_e$ (5 Experts) 
& 1.0000$_{\pm\text{0.00}}$    
& 1.0000$_{\pm\text{0.00}}$   
& 1.0000$_{\pm\text{0.00}}$   
& 1.0000$_{\pm\text{0.00}}$ \\ 

$D_e$ (10 Experts) 

& 0.9993$_{\pm\text{0.00}}$   
& 0.9809$_{\pm\text{0.03}}$ 
& 1.0000$_{\pm\text{0.00}}$    
& 1.0000$_{\pm\text{0.00}}$ \\

$D_e$ (15 Experts) 
& 0.9907 $_{\pm\text{0.01}}$    
& 0.9238$_{\pm\text{0.03}}$   
& 1.0000$_{\pm\text{0.00}}$    
& 1.0000$_{\pm\text{0.00}}$ \\

$D_e$ (20 Experts) 
& 0.9971 $_{\pm\text{0.01}}$    
& 0.9428$_{\pm\text{0.09}}$   
& 1.0000$_{\pm\text{0.00}}$    
& 1.0000$_{\pm\text{0.00}}$ \\

$D_e$ (25 Experts) 
& 0.9981 $_{\pm\text{0.01}}$    
& 0.9429$_{\pm\text{0.00}}$   
& 0.9999$_{\pm\text{0.00}}$    
& 0.9762$_{\pm\text{0.00}}$ \\

\bottomrule
\end{tabular}
\end{table}

\subsection{Implementation and results}
\begin{figure}[!ht]
    \centering
    \includegraphics[width=0.50\textwidth]{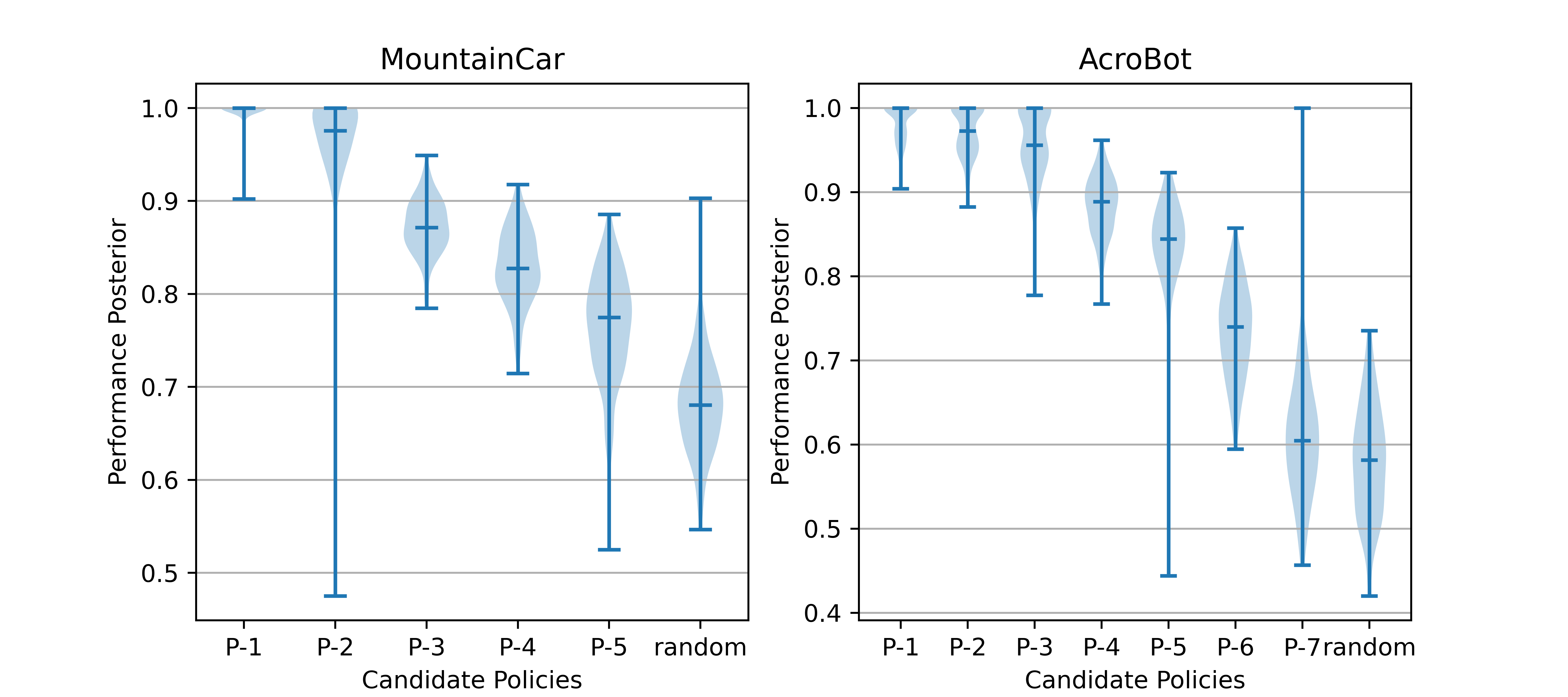}
    \caption{Posterior Comparison Between Candidate Policies Shows POPR-EABC is Effective in Estimating Performance and Helpful for Ranking.}
    \label{fig:posteriorsl}
\end{figure}

Based on the efficient estimation of POPR-EABC, we could conveniently extend the framework to multiple expert policies. As shown in Figure~\ref{fig:extension}, given $n$ expert policies (or more accurately, $n$
 expert-generated datasets), after sampling we have $m$ 
 datasets from $m$
 expert $\pi_{e_i}$, $i$ belongs to 1~$m$. For each candidate policy, e.g., policy $\pi_k$, we could conduct POPR-EABC posterior estimation instructed by each expert one, and then given by the posterior sample sets $\mathcal{S}_{\theta}^{k(i)}$, we could conduct action $A$ to understand each expert's evaluation. Since as long as exist a few supports from experts, this means the potential of $\pi_k$'s high performance. So we implemented a simple version of top-$r$ argmax posterior of probability $\theta$ and select for aggregation to $\mathcal{S}_{\theta}^{k}$
 from $r$ expert's opinions. The results are shown in Table~\ref{tab:extension} and Figure~\ref{fig:extension}, k = 3 in these experiments. We can see the candidate policies are distinguished well in Figure~\ref{fig:posteriorsl}.

